\pdfoutput=1
\documentclass[journal,twoside,web]{ieeecolor}
\usepackage{tmi}
\usepackage{cite}
\usepackage{amsmath,amssymb,amsfonts}
\usepackage{algorithmic}
\usepackage{graphicx}
\usepackage{textcomp}
\usepackage{booktabs}
\usepackage{multirow}
\usepackage{multicol}
\usepackage{array}
\usepackage[most]{tcolorbox}
\usepackage{xcolor}
\newcolumntype{C}[1]{>{\centering\arraybackslash}m{#1}}
\usepackage{adjustbox}
\usepackage{authblk}
\usepackage{url}
\usepackage{hyperref}
\def\BibTeX{{\rm B\kern-.05em{\sc i\kern-.025em b}\kern-.08em
    T\kern-.1667em\lower.7ex\hbox{E}\kern-.125emX}}
\begin{document}
\title{Generalised Medical Phrase Grounding}
\author{Wenjun~Zhang, Shekhar~S.~Chandra, and Aaron~Nicolson
\thanks{W. Zhang (e-mail: wenjun.zhang@uq.net.au) and S.~S.~Chandra are with The University of Queensland, Brisbane, Australia; W. Zhang and A.~Nicolson are with the Australian e-Health Research Centre, CSIRO Health and Biosecurity, Australia.}%
}
\maketitle
\begin{abstract}
Medical phrase grounding (MPG) maps textual descriptions of radiological findings to corresponding image regions. These grounded reports are easier to interpret, especially for non-experts. Existing MPG systems mostly follow the referring expression comprehension (REC) paradigm and return exactly one bounding box per phrase. Real reports often violate this assumption. They contain multi-region findings, non-diagnostic text, and non-groundable phrases, such as negations or descriptions of normal anatomy. Motivated by this, we reformulate the task as generalised medical phrase grounding (GMPG), where each sentence is mapped to zero, one, or multiple scored regions. To realise this formulation, we introduce the first GMPG model: MedGrounder. We adopted a two-stage training regime: pre-training on report sentence--anatomy box alignment datasets and fine-tuning on report sentence--human annotated box datasets. Experiments on PadChest-GR and MS-CXR show that MedGrounder achieves strong zero-shot transfer, state-of-the-art overall performance on PadChest-GR, competitive results on MS-CXR, and the largest gains on multi-region phrases, while using far fewer human box annotations. Finally, we show that MedGrounder can be composed with existing report generators to produce grounded reports without retraining the generator.
\end{abstract}

\section{Introduction}

Medical phrase grounding (MPG) maps textual descriptions of radiological findings to spatial regions in medical images. Presenting findings with their associated bounding boxes allows the text to be easier to interpret, especially by non-experts \cite{Zhang2025Anatomical, 10.1007/978-3-031-43990-2_35, TanidaInteractiveGeneration}.

\begin{figure}
    \centering
    \includegraphics[width=1\linewidth]{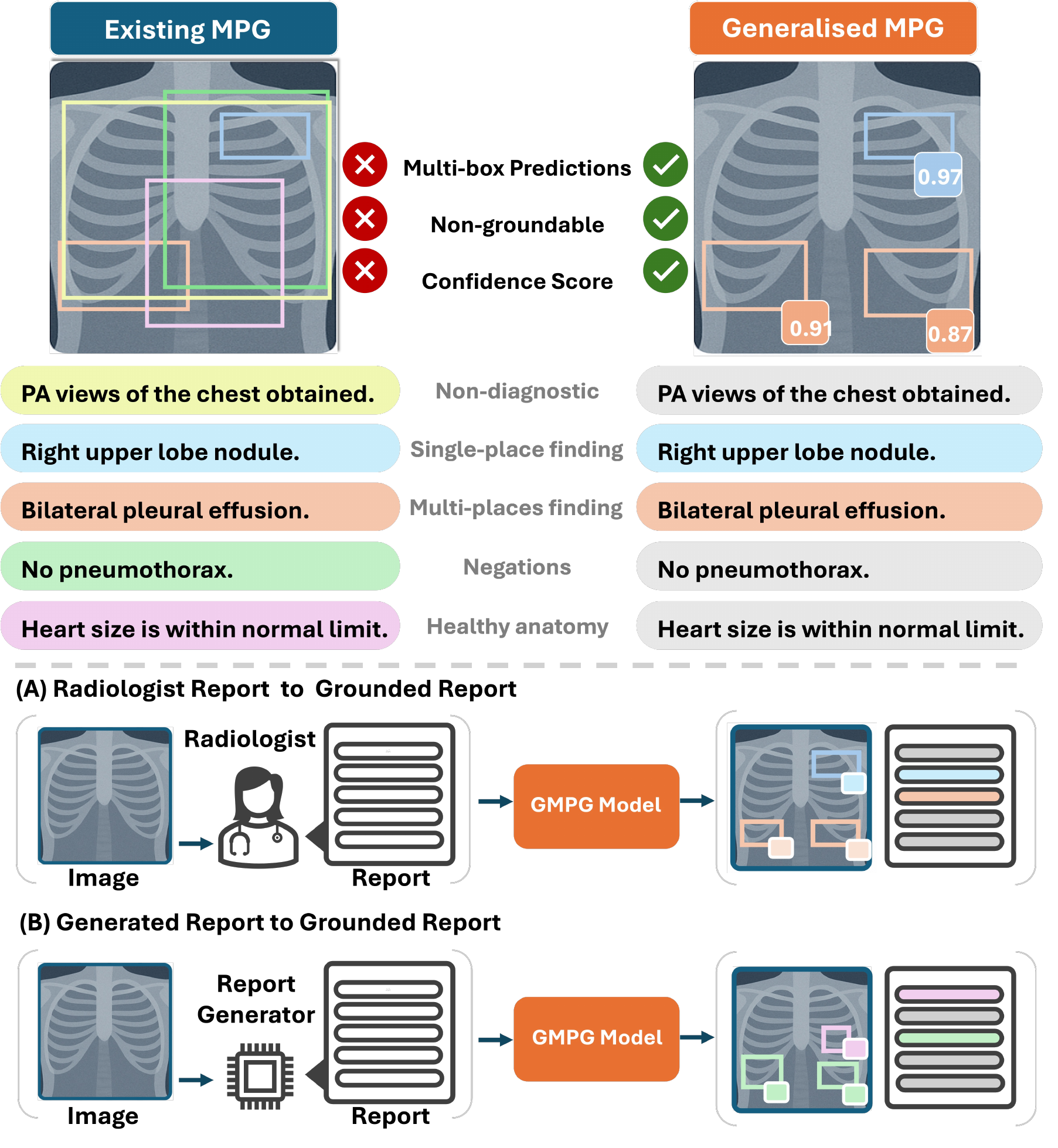}
    \caption{\textbf{Top}: MPG predicts a single bounding box per sentence, while GMPG additionally supports (i) multiple boxes per sentence, (ii) suppression of boxes for non-groundable phrases (e.g., negations), and (iii) confidence scoring for all predictions. \textbf{Bottom}: Applications of GMPG: (A) grounding a radiologist-written report for patient comprehension, and (B) grounding an AI-generated report for radiologist verification.}
    \label{fig:compare_app}
\end{figure}

Despite their promise, contemporary MPG systems exhibit several notable limitations. Most methods follow the referring expression comprehension (REC) framework. As shown in Figure \ref{fig:compare_app} (top left), the model takes an image and phrase as input and predicts a bounding box by regressing its coordinates, creating a one-to-one mapping between each phrase and region \cite{10.1007/978-3-031-43990-2_35, Zhang2025Anatomical}. This one-phrase/one-box assumption is mismatched to radiology reporting practice. Radiology reports contain findings corresponding to multiple regions (e.g., ``bilateral lower lobe opacities"), phrases that should not be grounded, including negated findings (e.g., ``No pneumothorax"), descriptions of normal anatomy, and non-diagnostic text like ``PA views provided". These cases are common in standard MPG benchmarks: in MS-CXR, multi-region phrases comprise approximately 23\% of phrases, and in PadChest-GR, over half of phrases are associated with zero or more regions. Existing REC-based MPG methods typically remove such phrases during training and evaluation, rather than modelling them explicitly, which limits their clinical usefulness \cite{10.1007/978-3-031-43990-2_35, Zhang2025Anatomical}. Moreover, the REC formulations return a box unconditionally and lack an explicit confidence signal to support abstention on non-groundable phrases or uncertainty when image evidence is insufficient. Confidence matters in clinical use because it is important to know when the model is uncertain.

To address these domain-specific requirements, we introduce \textbf{generalised medical phrase grounding (GMPG)}. GMPG reformulates MPG as a set prediction problem that supports zero-to-many grounded regions per phrase together with confidence scores. This formulation aligns with generalised referring expression comprehension (GREC), which treats grounding as set prediction in the general domain \cite{He2023GREC, Xiao2024VisualGroundingSurvey}. However, expert annotations are scarce and expensive. In preliminary experiments, directly fine-tuning a GREC-style set prediction model (MDETR) on expert-annotated MPG datasets yields near-zero localisation performance (Table~\ref{tab:two_stage_ablation}).

To make GMPG possible, we propose a two-stage weak-to-expert training pipeline. In Stage~1, we pre-train on \textbf{GMPG-ImaGenome}, a processed version of Chest ImaGenome \cite{WuChestReasoning}, where report sentences are linked to anatomical boxes produced by an anatomy detector. To convert this noisy structure into effective weak supervision for GMPG, we develop an LLM-assisted cleaning pipeline that removes redundant anatomical regions and filters spurious boxes for non-groundable sentences (removing approximately 80\% of boxes overall). In Stage~2, we fine-tune on expert-annotated datasets (MS-CXR \cite{BioViL_10.1007/978-3-031-20059-5_1} and PadChest-GR \cite{de_Castro_2025}). MedGrounder achieves state-of-the-art overall performance on PadChest-GR and competitive overall performance on MS-CXR, with the largest gains on multi-region phrases.

As the capabilities of GMPG differ from REC, standard REC metrics, such as single-box mIoU, are insufficient. Instead, we introduce new evaluation criteria suitable for the expanded task: (i) correct abstention on non-groundable phrases, (ii) strict exact set correctness for multi-region grounding (where missing or extra boxes are penalised), and (iii) localisation quality under both strict and boundary-tolerant criteria. Concretely, we report Precision@(F1$=$1, IoU$\geq$0.5) and Negative Accuracy (N-Acc) following \cite{He2023GREC}, as well as Center-Hit F1 (CH-F1) and Mask IoU accuracy to account for annotation variability (Section~\ref{sec:eval_metrics_gmpg}).

Beyond phrase grounding, MedGrounder enables a modular approach to grounded report generation (GRG) (Figure~\ref{fig:compare_app}, bottom). In GRG, a model generates a radiology report in which each described finding is accompanied by one or more bounding boxes localising it on the image \cite{bannur2024maira2}. By linking each finding to visual evidence, GRG can facilitate efficient radiologist verification of generated reports \cite{Yildirim2024MultimodalRadiology, bannur2024maira2}. Existing approaches train end-to-end grounded generators on expensive large-scale image--report--box triplets to do grounded report generation. In contrast, MedGrounder can be applied post-hoc to ground the outputs of any report generator, bypassing the annotation bottleneck while remaining compatible with state-of-the-art generation systems. We evaluate GRG using the RadFact framework from MAIRA-2 \cite{bannur2024maira2}, which uses an LLM to assess generated reports at the sentence level along three axes: whether each sentence is factually correct (logical F1), whether correct sentences are spatially supported (grounding F1), and overall spatial accuracy (spatial F1).

\subsubsection*{\textbf{Contributions}}
\begin{itemize}
    \item We formulate \textbf{Generalised Medical Phrase Grounding}, extending MPG to a clinically aligned set prediction task that supports zero, one, or multiple grounded regions per phrase with confidence scores, and propose an evaluation protocol for abstention and multi-region grounding.

    \item We release \textbf{GMPG-ImaGenome}, a weakly annotated pre-training resource derived from Chest ImaGenome via an LLM-assisted cleanup pipeline that removes redundant anatomical regions and filters spurious boxes for non-groundable sentences.

    \item We propose a \textbf{weak-to-expert training curriculum} for GMPG. Built on this curriculum, our \textbf{MedGrounder} (an MDETR-style model) achieves state-of-the-art overall performance on PadChest-GR and competitive overall performance on MS-CXR, with the largest gains on multi-region phrases.

    \item We demonstrate a \textbf{modular, data-efficient GRG strategy} by composing MedGrounder with existing report generators to produce grounded reports without requiring large-scale image--report--box supervision.

    \item Our code is available at \url{https://github.com/aehrc/MedGrounder} and the GMPG-ImaGenome will be released on PhysioNet.
\end{itemize}

\section{Related Work}
\begin{table*}[!htbp]
	\centering
	\caption{\textbf{Comparison of Prior Work Based on GMPG Requirements. }The table positions existing medical phrase grounding and related models against the three core requirements of GMPG: \textbf{Multi-region} (support for zero, one, or multiple regions per phrase), \textbf{Groundability} (ability to abstain from grounding), and \textbf{Confidence }(provision of confidence scores for predictions).}
	\label{tab:rw_matrix}
	\footnotesize
	\begin{tabular}{lllllll}
	\toprule
	\textbf{Method} & \textbf{Year-venue} & \textbf{Box supervision} & \textbf{Learning paradigm} & \textbf{Multi-region} & \textbf{Groundability} & \textbf{Confidence}\\
	\addlinespace[2pt]
    \midrule
    \addlinespace[2pt]
	GLoRIA & 2021-ICCV & No & Contrastive &  $\checkmark$ & $\times$ & $\times$  \\
	BioViL & 2022-ECCV & No & Contrastive & $\checkmark$  & $\times$ & $\times$  \\
	LDM & 2024-JBHI & No & Diffusion & $\checkmark$ & $\times$ & $\times$  \\
	Generate to Ground & 2025 MIDL & No & Diffusion & $\checkmark$ & $\times$ & $\times$ \\
    \midrule
	MedRPG & 2023-MICCAI & Yes & MPG / REC & $\times$ & $\times$ & $\times$  \\
	VG-CT & 2023-MICCAI & Yes & Multi-task &$\times$ & $\times$ & $\times$  \\
	ChEX & 2024-ECCV & Yes & Multi-task & $\checkmark$ & $\times$ & $\checkmark$ \\
	MedRG & 2024-CORR & Yes & Multi-task & $\times$ & $\times$ & $\checkmark$\\
	MAIRA-2 & 2024 & Yes & GRG & $\checkmark$ & $\checkmark$ & $\times$ \\
    AGPT & 2025-ISBI & Yes & MPG / REC & $\times$ & $\times$ & $\times$  \\
	\midrule

	\textbf{MedGrounder (ours)} & 2025 & Yes & GMPG / GREC & $\checkmark$ & $\checkmark$ & $\checkmark$ \\
	\bottomrule
	\end{tabular}
	\end{table*}

% In this section, we review existing literature across three relevant directions. First, we review MPG, categorised based on whether box supervision is used or not. Second, we examine the traditional REC framework, the source of the single-box limitation in MPG. Finally, we situate our work relative to the closely related task of GRG, which also aims to link text and visual evidence. These are summarised in Table \ref{tab:rw_matrix}.

We review literature across three domains: MPG, REC (and GREC), and GRG, with methods summarised in Table~\ref{tab:rw_matrix}.

\subsection{Medical phrase grounding (MPG)}
\subsubsection{Without box supervision}
Initial approaches to MPG did not rely on box supervision, learning local alignments from image-text pairs without explicit bounding box annotations. Contrastive learning models like GLoRIA \cite{GLoRIA_Huang_2021_ICCV}, BioViL \cite{BioViL_10.1007/978-3-031-20059-5_1} and BioViL-T \cite{10204115} exemplify this; they naturally produce heatmaps from their alignment scores. The MS-CXR benchmark \cite{BioViL_10.1007/978-3-031-20059-5_1} was, in fact, introduced to evaluate this emergent localisation capability against ground-truth boxes, rather than as a primary training set for grounding models. More recently, latent diffusion models \cite{vilouras2024zero, nutzel2025generate} have been used to extract cross-attention maps for zero-shot localisation. While these methods offer a way to learn alignments without expensive box annotations, their reliance on coarse heatmaps often limits localisation precision.

\subsubsection{With box supervision}
Methods trained with box supervision typically achieve higher precision than the previously mentioned non-box-supervised approaches. MedRPG pioneered this approach by formally defining the MPG task \cite{10.1007/978-3-031-43990-2_35}. It was also the first to use the MS-CXR benchmark for training explicit box predictors, rather than solely for evaluation. MedRPG adapted the REC-style TransVG model for medical images, introducing the TaCo module to strengthen region-phrase alignment. AGPT  extended this approach by adding an anatomical pre-training stage on anatomy name-box pairs before fine-tuning on expert-annotated sentence-level data \cite{Zhang2025Anatomical}. However, these methods all share a fundamental limitation: they adopt the traditional REC framework, which constrains them to predict exactly one bounding box per phrase. Consequently, they are trained on single-box subsets of datasets and cannot handle multi-region findings or non-groundable phrases.

Subsequent work has attempted to address parts of this limitation. MedRG \cite{ZouMedRG:Model} reframed the task as medical report grounding to handle non-groundable text. This multi-stage approach first uses a fine-tuned language model to extract a single, supposedly groundable phrase from a full report, which a separate decoder then grounds. While this adds a mechanism to filter some non-groundable phrases, it introduces architectural complexity and, critically, retains the single-box constraint, failing to support multi-region findings.

Beyond REC, other methods couple grounding with different objectives. ChEX \cite{ChEX_10.1007/978-3-031-72664-4_6} introduced a model based on DETR \cite{Carion2020End-to-EndTransformers} for interactive localisation and region description, which is capable of predicting multiple boxes with confidence scores. VG-CT \cite{Ichinose_2023} proposed a model that outputs segmentation masks from text, given an image and anatomical segmentation priors. However, none of the existing MPG methods satisfies all three core requirements for GMPG: multi-region support, groundability, and confidence scoring.

\subsection{Referring expression comprehension (REC) and its generalisation}
The single-box constraint observed in MPG stems from the adoption of the traditional REC framework from the general domain. Traditional REC tasks assume every textual expression refers to exactly one object in the image. This assumption is embedded in benchmark datasets like ReferIt and RefCOCO(g) \cite{kazemzadeh-etal-2014-referitgame, krishna2017visual, yu2016modeling}, which contain only groundable phrases with a single corresponding bounding box. Architecturally, REC-styled models regress directly on four bounding box coordinates without producing confidence scores, making single-box prediction intrinsic to their design.

This limitation led to GREC, which reframes grounding as set prediction: predict a variable-sized set of bounding boxes—including zero when appropriate \cite{He2023GREC, Hemanthage2024RECANTFormer}. Detection Transformers like MDETR  enable this by treating grounding as set prediction with cross-modal fusion and learnable object queries, naturally handling zero-or-more outputs with confidence scores \cite{Kamath2021MDETRUnderstanding}. Our work systematically adapts this more flexible formulation to medical imaging.

\subsection{Grounded report generation (GRG)}
Radiology report generation aims to automatically produce a free-text radiology report from an input image (or image series)\cite{NICOLSON2023102633, nicolson-etal-2025-impact, Liu_2021_CVPR, wu2025towards, zhou2025medversageneralistfoundationmodel, NICOLSON2024101585}. GRG extends this setting by producing a report while simultaneously localising the described findings \cite{bannur2024maira2, zhang2025development, luo-etal-2025-vividmed}. After each sentence is generated, the model typically outputs the corresponding bounding boxes that indicate the image regions supporting that sentence. GRG is closely related to GMPG, as it inherently involves phrase grounding as a subtask. Consequently, a GMPG model can be combined with an existing report generator to achieve GRG functionality, without the need for additional GRG-specific training.

MAIRA-2 \cite{bannur2024maira2} is the current state-of-the-art grounded report generation model we compare against. It accomplishes this by converting spatial coordinates into discrete spatial tokens. MAIRA-2 is trained in three modes aligned with three tasks: findings generation (FG), grounded report (GR), and phrase grounding (PG). The authors build task-specific instruction prompts and train each mode on different datasets. Notably, MAIRA-2 relies on training on large-scale, private datasets of image-sentence-box annotations.

\section{Methodology}\label{sec:methodology}
\begin{figure*}
    \centering
    \includegraphics[width=0.8\linewidth]{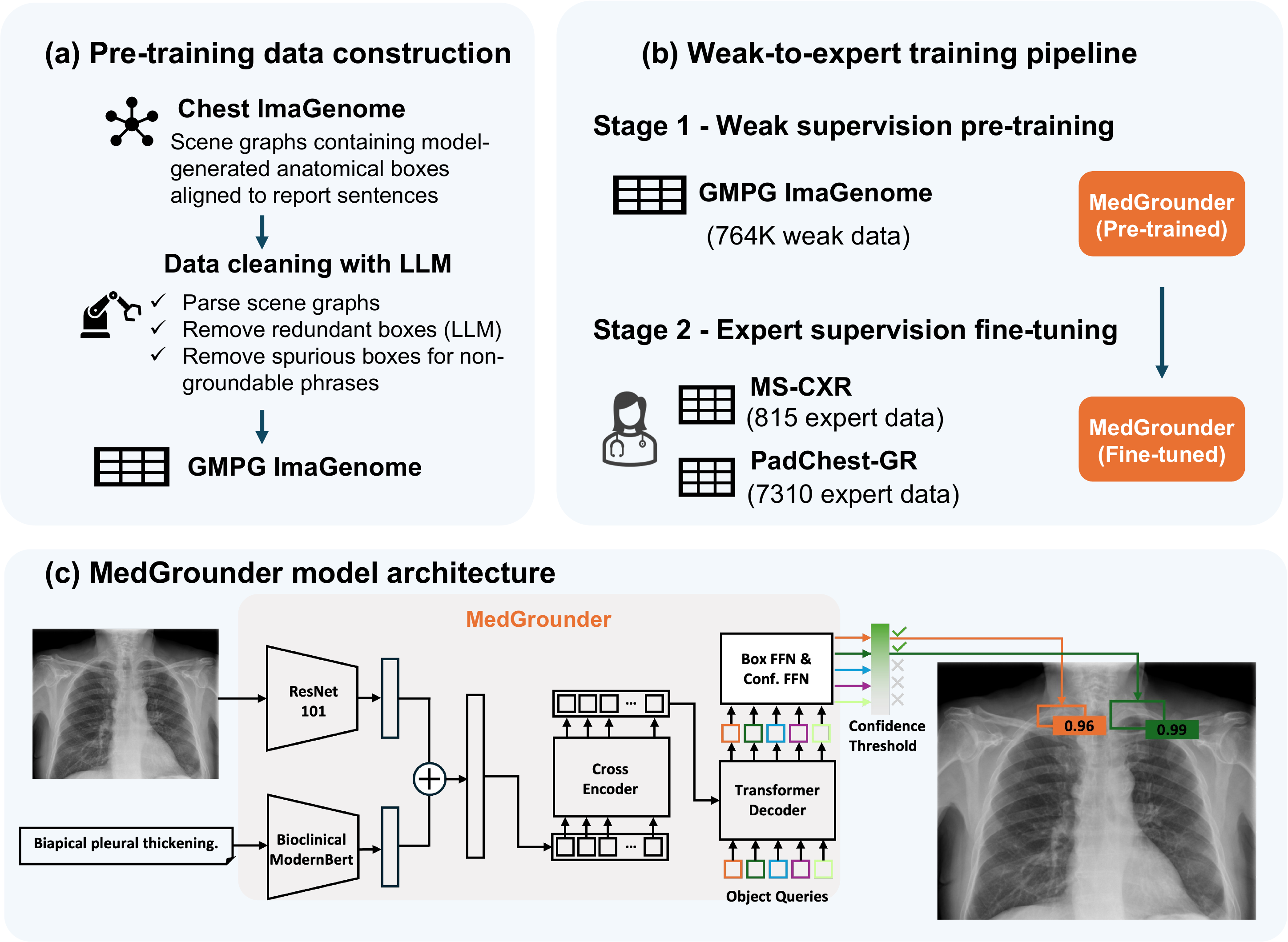}
    \caption{\textbf{MedGrounder framework overview.} (a) LLM-based data cleaning pipeline for constructing GMPG-ImaGenome. (b) Weak-to-expert two-stage training strategy. (c) Set-prediction architecture that outputs zero, one, or multiple scored bounding boxes per phrase.}
    \label{fig:framework}
\end{figure*}

Figure~\ref{fig:framework} summarises our approach. We first define GMPG, then construct GMPG-ImaGenome via an LLM-assisted cleaning pipeline, and finally train MedGrounder with a weak-to-expert curriculum using an MDETR-based set-prediction model.
\subsection{Generalised medical phrase grounding}

We define GMPG as mapping a medical image and a phrase to a \emph{set of regions}. Let the dataset be:
\begin{equation}\label{equ:dataset}
\mathcal{D} = \bigl\{(I_m, \mathcal{P}_m, \{S_{mi}\}_{i=1}^{N_m})\bigr\}_{m=1}^{M},
\end{equation}
consisting of $M$ image–report pairs, where $I_m$ denotes the image and $\mathcal{P}_m=\{p_{mi}\}_{i=1}^{N_m}$ denotes its report, consisting of $N_{m}$ sentences. $\{S_{mi}\}_{i=1}^{N_m}$ represents the ground-truth bounding-box sets, defined as $S_{mi}=\{ b_{mij} \}_{j=1}^{K_{mi}}$, where $b_{mij}\in[0,1]^4$ is a normalised box and $K_{mi}\ge 0$ is the region count for the sentence. GMPG maps a pair $(I,p)$ to a predicted set $\hat{S}_{mi} = f(I_m, p_{mi})$ approximating $S_{mi}$. The variable cardinality $K_{mi}$ supports multi-box grounding and non-groundable phrases, expanding upon MPG which assumes $|S_{mi}|=1$.

\subsection{Weak-to-expert training curriculum}

\subsubsection{Motivation}
Phrase grounding for chest X-rays remains data-limited compared with general-domain referring expression comprehension.
MS-CXR is a commonly used benchmark for medical phrase grounding. In preliminary experiments, directly adapting a general-domain grounding model (MDETR) to MS-CXR performs poorly, yielding near-zero localisation accuracy (Table~\ref{tab:two_stage_ablation}).
This motivates pre-training on a domain-specific dataset to improve downstream task performance.

\subsubsection{Weak supervision from Chest ImaGenome}
We identify Chest ImaGenome as a suitable source of weak supervision. It provides scene graphs over 29 anatomical regions with $1\,256$ anatomy--attribute relation types, and includes large-scale localised comparison relations across sequential exams. Importantly for our use case, their rule-based NLP pipeline associates report sentences with one or more anatomical regions, enabling sentence--region alignment at scale.

However, Chest ImaGenome is not designed for GMPG. Its sentence-to-region alignment exhibits two key limitations for the GMPG task:
(i) sentences are often linked to both coarse parent regions and fine-grained child regions (e.g., ``left lung'' and ``left lower lung zone''), introducing redundant parent boxes;
(ii) sentences describing normal findings or negated findings may still be linked to anatomical boxes, whereas GMPG requires \emph{zero-box} targets for these non-groundable phrases.
Figure~\ref{fig:gmpg_imagenome_case} illustrates this issue: the original Chest ImaGenome alignment can yield many boxes for reports dominated by normal/negated statements.

\subsubsection{Cleaning Chest ImaGenome for GMPG}
\begin{figure}
    \centering
    \includegraphics[width=1\linewidth]{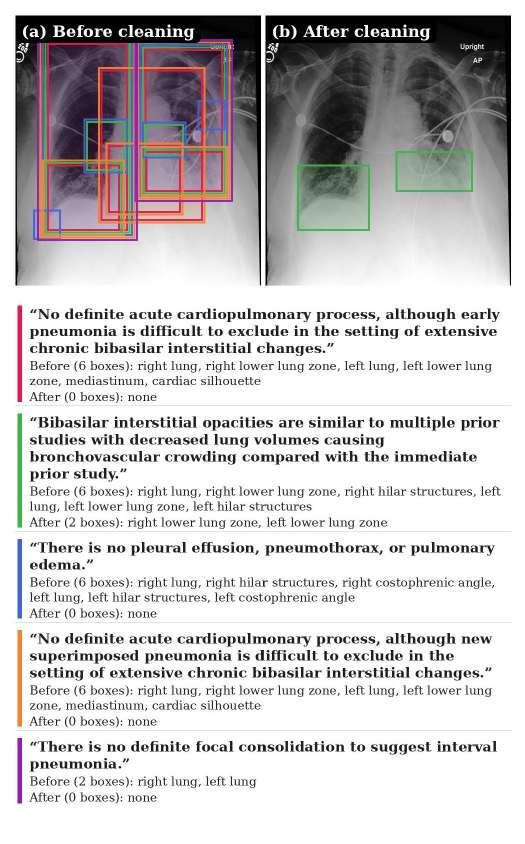}
    \caption{\textbf{Cleanup example.} Before cleaning, sentences are aligned to many (often redundant) anatomy boxes, including negated/normal statements; after cleaning, negated/normal sentences become zero-box targets and remaining boxes are filtered to specific regions.}
    \label{fig:gmpg_imagenome_case}
\end{figure}

\begin{figure}[t] % Use [t] to float to the top, saving space
\centering
% --- Use \footnotesize for all content in the figure ---
\footnotesize
% --- Combine all 3 boxes into ONE tcolorbox ---
% This saves all the vertical margin space between the 3 original boxes
\begin{tcolorbox}[
    title={LLM Prompt for Region Filtering}, % Single, descriptive title
    sharp corners,
    fonttitle=\bfseries, % Title font (will be \footnotesize)
    top=2pt, bottom=3pt, left=4pt, right=4pt, boxsep=2pt % Tight padding
]
 1. System Prompt
\textbf{System:} Given a radiology report sentence and its candidate anatomical regions, return only the most specific regions that correspond to the described abnormality. Remove broader or redundant regions. Output valid JSON with the key ``selected\_region'' only.

% --- Use a subtle separator ---
\par\smallskip\hrule\smallskip

% 2. User Prompt
\textbf{User:}
Sentence: ``There is minimal upper zone vascular redistribution without overt pulmonary edema.''

Candidates:
[``right lung'', ``right upper lung zone'', ``right hilar structures'',
 ``left lung'', ``left upper lung zone'', ``left hilar structures'']

% --- Use a subtle separator ---
\par\smallskip\hrule\smallskip

% 3. Expected Output
\textbf{Expected output:}
\{``selected\_regions'': [``right upper lung zone'', ``left upper lung zone'']\}
\end{tcolorbox}

\caption{\label{fig:prompt}\textbf{LLM prompt for filtering redundant anatomical regions.} The model selects the most specific regions from candidates, discarding broader parent regions.}

\vspace{-12pt}
\end{figure}

\begin{figure}
    \centering
    \includegraphics[width=1\linewidth]{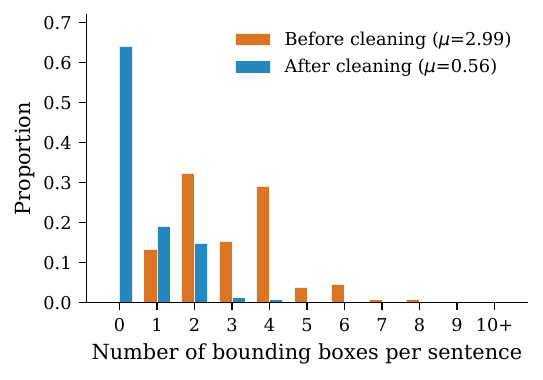}
    \caption{\textbf{Boxes per sentence.} Distribution before vs.\ after cleaning; the mean drops from $\mu=2.99$ to $\mu=0.56$, with many more zero-box sentences.}
    \label{fig:gmpg_imagenome_box_count}
\end{figure}

\begin{figure}
    \centering
    \includegraphics[width=1\linewidth]{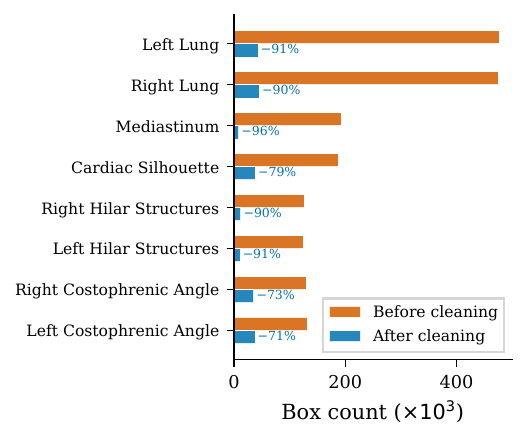}
    \caption{\textbf{Largest reductions by region.} Box counts drop most for coarse parent regions (e.g., left/right lung), indicating removal of redundant broad boxes.}
    \label{fig:gmpg_imagenome_region}
\end{figure}

To better match GMPG requirements, we transform each scene graph from Chest ImaGenome into a tabular format containing: the image, report sentence, anatomical structure name, anatomical bounding box, and a sentence-level normality label.
We then apply a two-stage cleaning procedure. First, we remove boxes associated with non-groundable sentences using the sentence-level normality label provided with the dataset. Second, for the remaining sentences, we perform region filtering to reduce parent--child redundancy by selecting anatomically appropriate regions conditioned on the sentence text and the candidate regions; we implement this filtering using LLaMA-3-70B \cite{meta2024llama3blog}, using the prompt in Figure~\ref{fig:prompt}.

After cleaning, the example in Figure~\ref{fig:gmpg_imagenome_case} reduces from 26 boxes across five sentences to a single sentence aligned to two boxes.
Figure~\ref{fig:gmpg_imagenome_box_count} shows the distribution of boxes per sentence before and after cleaning.
Overall, the cleaning reduces the box count by approximately 80\% (from 2\,400\,524 to 426\,749), with the largest reductions in coarse regions such as ``left lung'' and ``right lung'' (Figure~\ref{fig:gmpg_imagenome_region}), consistent with the parent-box redundancy.

\subsubsection{Weak-to-expert training}
The resulting dataset better fits GMPG by providing cleaner region supervision and reasonable zero-box targets for non-groundable sentences.
We use this cleaned Chest ImaGenome-derived data for the \emph{weak} pre-training stage: weak because the supervision aligns sentences to \emph{anatomical} regions (not finding extents), and the anatomical boxes are produced by an automated anatomy detector.
We then fine-tune the pre-trained model on expert-annotated phrase grounding benchmarks (e.g., MS-CXR and PadChest-GR), forming a weak-to-expert curriculum.
We release our cleaned dataset on PhysioNet, including additional annotations such as sentence-level finding classes, sentence-level normality labels, and image-level normality labels for potential reuse by future work.

\subsubsection{Datasets}
Our experiments utilise three chest X-ray datasets, summarised in Table~\ref{tab:dataset_stats}.

\textbf{GMPG-ImaGenome}, derived from Chest ImaGenome~\cite{WuChestReasoning} via the cleaning procedure described above, serves as weak pre-training data. It provides large-scale sentence--anatomy box pairs across 29 predefined regions, with rich variability in disease types and sentence structures. After cleaning, 64\% of sentence entries are zero-box (non-groundable), 19\% single-box, and 17\% multi-box, with a mean sentence length of $9.2$ words.

\textbf{PadChest-GR}~\cite{de_Castro_2025} and \textbf{MS-CXR}~\cite{BioViL_10.1007/978-3-031-20059-5_1} are smaller, expert-annotated datasets used for fine-tuning and evaluation. Both contain shorter, more focused text spans grounded to finding extents rather than anatomical regions. PadChest-GR comprises English translations of Spanish report sentences and includes zero-box cases (41\%), while MS-CXR consists of concise phrases extracted from approximately one thousand MIMIC-CXR~\cite{johnson2019mimiccxr} reports with no zero-box cases. Multi-box rates are 12--14\% in PadChest-GR and 22--24\% in MS-CXR. PadChest-GR provides full grounded reports, where each report sentence is annotated with zero or more bounding boxes. We decompose each report into individual sentences and evaluate grounding at the sentence level, consistent with our GMPG formulation.

\begin{table*}
\centering
\caption{\textbf{Statistics of Datasets for pre-training and Fine-tuning of MedGrounder.} The table details the purpose, splits, sizes, and annotation characteristics of Chest ImaGenome, PadChest-GR, and MS-CXR.}
\label{tab:dataset_stats}
% \setlength{\tabcolsep}{2.75pt}
% \begin{tabular}{p{60pt}p{40pt}p{60pt}p{40pt}p{20pt}C{30pt}C{50pt}C{30pt}C{80pt}C{45pt}}
\begin{tabular}{llllllllll}
\toprule
\textbf{Dataset} & \textbf{Purpose} & \shortstack{\textbf{Annotation}\\\textbf{source}} &
\shortstack{\textbf{Region}\\\textbf{type}} & \textbf{Split} & \textbf{\#Image} & \shortstack{\textbf{\#Image-}\\\textbf{phrase}} & \textbf{\#Boxes} & \shortstack{\textbf{\#Boxes per}\\\textbf{phrase (\%)}} &
\shortstack{\textbf{Sent. length}\\\textbf{Mean $\pm$ SD}} \\
\midrule
GMPG-ImaGenome & Pre-training & Model generated & Anatomy & train & $137\,065$ & $764\,540$ & $426\,749$ & 64 / 19 / 17 & 9.2 $\pm$ 5.8 \\
\midrule
\multirow{3}{*}{PadChest-GR} & \multirow{3}{*}{Fine-tuning} & \multirow{3}{*}{Human labeled} & \multirow{3}{*}{Finding} & train & $3\,185$ & $7\,310$ & $5\,404$ & 41 / 46 / 13 & 5.1 $\pm$ 3.2 \\
& & & & val & 455 & $1\,052$ & 808 & 39 / 46 / 14 & 5.2 $\pm$ 3.6 \\
& & & & test & 915 & $2\,112$ & $1\,521$ & 41 / 47 / 12 & 5.3 $\pm$ 3.5 \\
\midrule
\multirow{3}{*}{MS-CXR} & \multirow{3}{*}{Fine-tuning} & \multirow{3}{*}{Human labeled} & \multirow{3}{*}{Finding} & train & 737 & 815 & $1\,020$ & 0 / 77 / 23 & 5.7 $\pm$ 3.5 \\
& & & & val & 155 & 169 & 212 & 0 / 76 / 24 & 5.9 $\pm$ 3.4 \\
& & & & test & 155 & 176 & 216 & 0 / 78 / 22 & 5.5 $\pm$ 3.3 \\
\bottomrule
\end{tabular}
\end{table*}

\subsection{MedGrounder framework}

\subsubsection{Architecture}
MedGrounder adapts the MDETR framework~\cite{Kamath2021MDETRUnderstanding} for the GMPG setting. The image $I$ is encoded by a ResNet-101~\cite{He2016DeepResidual} backbone initialised from MDETR's checkpoints. The phrase $p$ is encoded by BioClinical ModernBERT~\cite{sounack2025bioclinicalmodernbertstateoftheartlongcontext}, an in-domain clinical text encoder selected on the basis of our ablation study (Table~\ref{tab:ablation_lang_all_plus_deltas}). A cross-modal encoder with bidirectional attention fuses the concatenated visual and textual features, which a Transformer decoder then attends to via $N_Q{=}5$ learnable object queries. Each query produces two outputs: a bounding box via an MLP FFN and a confidence score via a linear projection. This design naturally supports zero, one, or multiple active predictions per phrase. The visual encoder, cross-modal encoder, and Transformer decoder were all warm-started from the MDETR checkpoint. The Box and Confidence FFNs were randomly initialised rather than retaining MDETR weights, because the target pathology differs substantially from the general-domain objects.

\subsubsection{Hungarian matching and training objectives}
\label{sec:matching}
For each image--sentence pair $(I_m, p_{mi})$ with $K$ ground-truth
boxes $\{b_k\}_{k=1}^{K}$, the model predicts
$\hat{Y} = \{(\hat{b}_{j}, \hat{c}_{j})\}_{j=1}^{N_Q}$
(indices $m,i$ dropped for clarity).
Because $K \le N_Q$, we seek a one-to-one assignment
$\hat{\pi}:\{1,\dots,K\}\!\to\!\{1,\dots,N_Q\}$
that minimises the total matching cost:
\begin{equation}
\begin{split}
\hat{\pi} = \arg\min_{\pi} \sum_{k=1}^{K}
\Bigl[
  &-\hat{c}_{\pi(k)}
  + \lambda_{L1}\|\hat{b}_{\pi(k)} - b_{k}\|_1 \\
  &+ \lambda_{\mathrm{GIoU}}\bigl(1 - \mathrm{GIoU}(\hat{b}_{\pi(k)}, b_{k})\bigr)
\Bigr],
\end{split}
\label{eq:match}
\end{equation}
where $\hat{c}_{\pi(k)}$ is the predicted groundability confidence of the $\pi(k)$-th query, $\|\hat{b}_{\pi(k)} - b_k\|_1$ is the L1 distance between the predicted and ground-truth box coordinates, and $\mathrm{GIoU}(\hat{b}_{\pi(k)}, b_k)$ is the generalised intersection-over-union between the two box coordinates~\cite{Rezatofighi2019GeneralizedMetric}.
$\lambda_{L1}$ and $\lambda_{\mathrm{GIoU}}$ are loss weights.
This assignment is solved exactly by the Hungarian
algorithm~\cite{Kuhn1955Hungarian}.
After matching, we construct ground-truth labels $\mathbf{t} \in \{0,1\}^{N_Q}$, where $t_j = 1$ if $j = \hat{\pi}(k)$ for some $k$, and $t_j = 0$ otherwise. The Confidence FFN predicts confidence scores $\hat{c}_j$ for each query, and is supervised by $\mathbf{t}$ via the classification loss in Eq.~\eqref{eq:cls}.

The three GMPG grounding scenarios are handled as follows. For non-groundable phrases ($K=0$), there are no ground-truth boxes, so the matching step is skipped. All $N_Q$ queries receive target $t_j = 0$, supervising the model to predict low confidence for every query. For single-box cases ($K=1$), exactly one query---the one whose predicted box minimises Eq.~\eqref{eq:match}---is assigned $t_j = 1$; the remaining $N_Q - 1$ queries receive $t_j = 0$. For the multi-box cases ($K>1$), the Hungarian algorithm assigns a distinct query to each ground-truth box, so $K$ queries receive $t_j = 1$ and $N_Q - K$ receive $t_j = 0$. This is the key departure from REC-style models, which always and only predict four coordinates.

The groundability and localisation losses are:
\begin{align}
\mathcal{L}_{\mathrm{cls}}
  &= \frac{1}{N_Q}\sum_{j=1}^{N_Q}
     \alpha_{t_j}\,\mathrm{BCE}(\hat{c}_j, t_j),
  \label{eq:cls} \\
\mathcal{L}_{\mathrm{box}}
  &= \frac{1}{\max(1,K)}\sum_{k=1}^{K}
     \Bigl(
       \lambda_{L1}\|\hat{b}_{\hat{\pi}(k)} - b_k\|_1 \notag\\
  &\quad + \lambda_{\mathrm{GIoU}}\bigl[1 - \mathrm{GIoU}(
           \hat{b}_{\hat{\pi}(k)}, b_k)\bigr]
     \Bigr),
  \label{eq:box}
\end{align}
with total loss $\mathcal{L} = \mathcal{L}_{\mathrm{cls}} +
\mathcal{L}_{\mathrm{box}}$.
We set $\alpha_{t_j=1} = 1.0$ and $\alpha_{t_j=0} = 0.1$
to down-weight the abundant unmatched queries, following the
imbalance-reweighting strategy of DETR~\cite{Carion2020End-to-EndTransformers}.
The regression weights $\lambda_{L1}=5$ and $\lambda_{\mathrm{GIoU}}=2$
are adopted directly from DETR.
Note that $\mathcal{L}_{\mathrm{box}}$ contributes only when
$K \ge 1$; for non-groundable phrases ($K=0$) the denominator
becomes $\max(1,0)=1$ but the sum is empty, so only
$\mathcal{L}_{\mathrm{cls}}$ is active.

\subsection{Evaluation metrics for GMPG}
\label{sec:eval_metrics_gmpg}

Standard REC metrics (e.g., mIoU) assume a single predicted box per phrase and cannot penalise over- or under-prediction. We report four metrics that cover the three GMPG requirements.

\textbf{Negative Accuracy (N-Acc)} is the proportion of
non-groundable phrases ($K{=}0$) for which the model correctly
produces no box. It directly measures safe abstention on negated or normal findings.

\textbf{Precision@(F$_1${=}1, IoU$\geq$0.5)} counts a phrase as correct only if the predicted set exactly matches the ground truth under one-to-one matching with IoU$\geq$0.5, penalising both missing and extra boxes. It is the primary metric for set correctness.

\textbf{Center-Hit F1 (CH-F1)} relaxes boundary sensitivity by requiring only that each predicted box centre falls inside the ground-truth region, then reports F1 over precision and recall. Strict boundary agreement is often unreliable for diffuse findings (e.g., consolidation, atelectasis) whose extent is inherently ambiguous, and radiologists vary considerably in how tightly or loosely they draw bounding boxes around the same finding. CH-F1 therefore measures whether the model finds the right region, regardless of how the box edges are drawn.

\textbf{Mask IoU Accuracy} converts all predicted and ground-truth boxes for a phrase into filled binary masks (i.e., pixels inside each box are set to 1) and merges them into a single unified mask, then reports the proportion of phrases whose mask IoU$\geq$0.5. This mask-level merging naturally handles multi-box phrases without requiring one-to-one box correspondence.

Together, these four metrics probe complementary failure modes: P@F1=1 penalises both missing boxes and spurious extra boxes, making it the strictest test of set-level localisation; N-Acc captures whether the model safely abstains on non-groundable phrases; CH-F1 measures whether predicted boxes target the correct region even when boundary delineation is imprecise; and Mask IoU Accuracy evaluates boundary precision regardless of how many boxes the model predicts.

\subsection{Evaluation metrics for grounded report generation}
For grounded report generation, we adopt the RadFact evaluation framework introduced in MAIRA-2~\cite{bannur2024maira2}. RadFact leverages large language models to assess report quality at the sentence level through three hierarchical metrics: (i)~\textit{logical F1} measures whether generated sentences are factually entailed by the ground-truth report; (ii)~\textit{grounding F1} evaluates whether logically correct sentences are also spatially entailed; and (iii)~\textit{spatial F1} measures the fraction of all grounded sentences that are both logically and spatially correct.

\begin{figure*}[!htbp]
    \centering
    \includegraphics[width=1\linewidth]{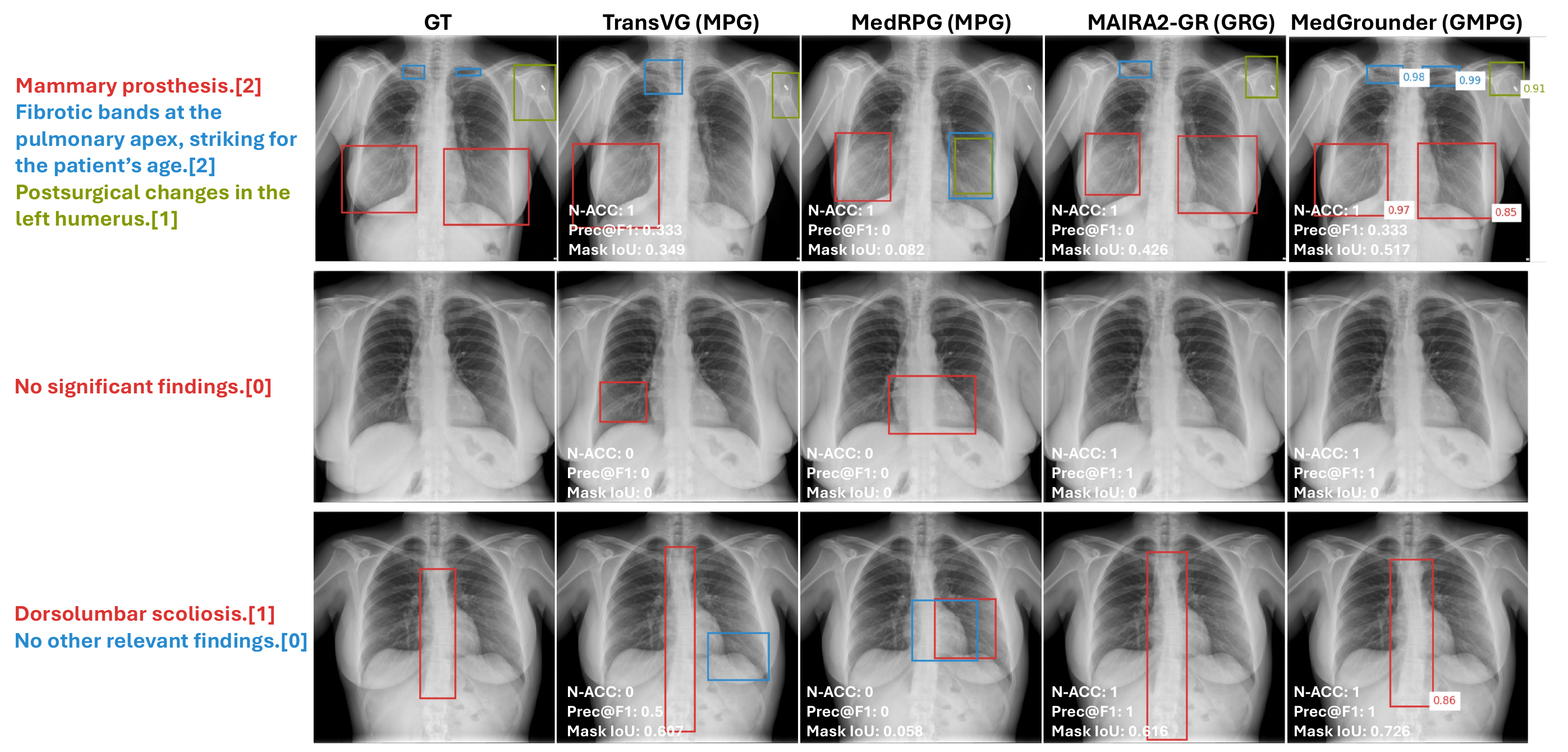}
    \caption{\textbf{Qualitative comparison of grounding results from TransVG, MedRPG, MAIRA-2 GR, and the proposed MedGrounder on three PadChest-GR examples.} Ground-truth (GT) is shown in the first column, model predictions in the remaining columns. Numbers in brackets after each sentence indicate the number of bounding boxes in the GT.
}
        \label{fig:mpg_gmpg}
\end{figure*}

\section{Experiment setup}
\subsubsection{Data preparation} Following MedRPG, each image was resized so that its longer side was 640 pixels while preserving aspect ratio, then zero-padded to a $640\times640$ square. Training time image augmentations included Gaussian noise and random cropping. Images were normalised with ImageNet mean and standard deviation, following DETR. Boxes were transformed with the images, and any with less than 90\% visibility after augmentation were discarded.

\subsubsection{Training and inference setup} MedGrounder was first pre-trained on the cleaned Chest ImaGenome for 15 epochs (batch size 32) using AdamW \cite{DBLP:conf/iclr/LoshchilovH19} with a 1e-5 learning rate; the checkpoint with the best validation Mask IoU was kept for fine-tuning. In the second stage, we fine-tuned on PadChest-GR and MS-CXR for 15 epochs, selecting the final model by the much stricter metric $P@F_1{=}1$. We performed five runs and computed the mean and standard deviation. All runs used a single NVIDIA H100 96GB GPU. At inference time, predictions were filtered by an optimised confidence threshold at 0.8 and merged with weighted box fusion (WBF).

\subsubsection{Baselines}
We compared MedGrounder against state-of-the-art models in two distinct tasks.
\paragraph{GMPG}
We benchmarked against two families of box-supervised baselines. The first was REC-style models, which included TransVG, MedRPG, and AGPT. The second was GRG models, for which we compared against MAIRA-2. To ensure a fair comparison, all models were evaluated on the official MS-CXR split; MedRPG and AGPT were retrained accordingly. Furthermore, MAIRA-2 was evaluated in both phrase-grounding (PG) and grounded-report (GR) modes, as its original training was split by mode (PadChest-GR for GR and MS-CXR for PG).

\paragraph{GRG}
To evaluate MedGrounder for GRG, we compared an end-to-end grounded report generation model (MAIRA-2 GR) with two pipeline approaches: MAIRA-2 FG + MedGrounder and CXRMate-RRG24 \cite{nicolson2024ehealthcsirorrg24entropyaugmented} + MedGrounder. Because the training data for CXRMate-RRG24 (the original PadChest) overlaps with the PadChest-GR test set, we created a 38-case non-overlapping subset to ensure a fair evaluation of all three models.
% ============================================================
% TABLE 1: PadChest-GR
% ============================================================
\begin{table*}[t]
\centering
\caption{\textbf{Phrase grounding performance on the test set of PadChest-GR.} Models are grouped into (A)~not trained on PadChest-GR and (B)~trained/fine-tuned on PadChest-GR. Best results are \textbf{bold}, second best are \underline{underlined}.}
\label{tab:padchest_results}
\resizebox{\textwidth}{!}{%
\begin{tabular}{ll cc c ccc ccc ccc}
\toprule
& & & & \textbf{No} & \multicolumn{3}{c}{\textbf{Single ($n{=}983$)}} & \multicolumn{3}{c}{\textbf{Multi ($n{=}255$)}} & \multicolumn{3}{c}{\textbf{Overall ($n{=}2{,}112$)}} \\
\cmidrule(lr){5-5} \cmidrule(lr){6-8} \cmidrule(lr){9-11} \cmidrule(lr){12-14}
\textbf{Model} & \textbf{Fw.} & \textbf{Data} & \textbf{Ann.}
& $\textbf{N-Acc}\phantom{_{\pm0.0}}$ & $\textbf{P@1}\phantom{_{\pm0.0}}$ & $\textbf{CH-F1}\phantom{_{\pm0.0}}$ & $\textbf{Acc}\phantom{_{\pm0.0}}$ & $\textbf{P@1}\phantom{_{\pm0.0}}$ & $\textbf{CH-F1}\phantom{_{\pm0.0}}$ & $\textbf{Acc}\phantom{_{\pm0.0}}$ & $\textbf{P@1}\phantom{_{\pm0.0}}$ & $\textbf{CH-F1}\phantom{_{\pm0.0}}$ & $\textbf{Acc}\phantom{_{\pm0.0}}$ \\
\midrule
\multicolumn{14}{l}{\textit{(A) Models not trained on PadChest-GR}} \\[2pt]
MedRPG              & MPG  & M*  & 624   & $\underline{0}\phantom{_{\pm0.0}}$    & $\underline{13.2}\phantom{_{\pm0.0}}$ & $35.4\phantom{_{\pm0.0}}$          & $12.7\phantom{_{\pm0.0}}$          & $\underline{0}\phantom{_{\pm0.0}}$   & $34.9\phantom{_{\pm0.0}}$          & $\underline{0.8}\phantom{_{\pm0.0}}$ & $\underline{6.2}\phantom{_{\pm0.0}}$ & $25.9\phantom{_{\pm0.0}}$          & $\underline{10.3}\phantom{_{\pm0.0}}$ \\
AGPT (pre.)         & MPG  & --  & 0     & $\underline{0}\phantom{_{\pm0.0}}$    & $13.0\phantom{_{\pm0.0}}$          & $\underline{39.9}\phantom{_{\pm0.0}}$ & $\underline{12.9}\phantom{_{\pm0.0}}$ & $\underline{0}\phantom{_{\pm0.0}}$   & $\underline{37.4}\phantom{_{\pm0.0}}$ & $0\phantom{_{\pm0.0}}$            & $6.1\phantom{_{\pm0.0}}$          & $\underline{29.6}\phantom{_{\pm0.0}}$ & $\underline{10.3}\phantom{_{\pm0.0}}$ \\
MedGrounder (pre.)  & GMPG & --  & 0     & $\textbf{86.5}\phantom{_{\pm0.0}}$    & $\textbf{18.1}\phantom{_{\pm0.0}}$ & $\textbf{60.2}\phantom{_{\pm0.0}}$ & $\textbf{21.2}\phantom{_{\pm0.0}}$ & $\textbf{8.2}\phantom{_{\pm0.0}}$    & $\textbf{65.3}\phantom{_{\pm0.0}}$ & $\textbf{16.1}\phantom{_{\pm0.0}}$ & $\textbf{45.2}\phantom{_{\pm0.0}}$ & $\textbf{58.2}\phantom{_{\pm0.0}}$ & $\textbf{20.1}\phantom{_{\pm0.0}}$ \\
\midrule
\multicolumn{14}{l}{\textit{(B) Models trained on PadChest-GR}} \\[2pt]
TransVG             & MPG  & M*,P* & 3\,993  & $0\phantom{_{\pm0.0}}$              & $\textbf{54.4}\phantom{_{\pm0.0}}$ & $\textbf{82.6}\phantom{_{\pm0.0}}$ & $\textbf{54.2}\phantom{_{\pm0.0}}$ & $0\phantom{_{\pm0.0}}$               & $56.8\phantom{_{\pm0.0}}$          & $5.1\phantom{_{\pm0.0}}$           & $25.3\phantom{_{\pm0.0}}$          & $56.0\phantom{_{\pm0.0}}$          & $44.1\phantom{_{\pm0.0}}$ \\
MAIRA-2 PG          & GRG  & U,M,P & 73\,043 & $72.1\phantom{_{\pm0.0}}$           & $39.6\phantom{_{\pm0.0}}$          & $74.4\phantom{_{\pm0.0}}$          & $40.8\phantom{_{\pm0.0}}$          & $14.9\phantom{_{\pm0.0}}$            & $73.1\phantom{_{\pm0.0}}$          & $26.7\phantom{_{\pm0.0}}$          & $50.0\phantom{_{\pm0.0}}$          & $66.1\phantom{_{\pm0.0}}$          & $37.9\phantom{_{\pm0.0}}$ \\
MAIRA-2 GR          & GRG  & U,M,P & 73\,043 & $\textbf{91.0}\phantom{_{\pm0.0}}$  & $45.8\phantom{_{\pm0.0}}$          & $\underline{79.5}\phantom{_{\pm0.0}}$ & $47.0\phantom{_{\pm0.0}}$       & $14.9\phantom{_{\pm0.0}}$            & $74.1\phantom{_{\pm0.0}}$          & $31.8\phantom{_{\pm0.0}}$          & $60.7\phantom{_{\pm0.0}}$          & $75.1\phantom{_{\pm0.0}}$          & $43.9\phantom{_{\pm0.0}}$ \\
\textbf{MedGrounder} & GMPG & P    & 7\,315  & $89.7_{\pm1.1}$ & $\underline{47.6}_{\pm0.3}$ & $78.0_{\pm0.6}$ & $\underline{49.2}_{\pm0.5}$ & $\underline{24.5}_{\pm1.9}$ & $\underline{77.6}_{\pm0.7}$ & $\underline{36.8}_{\pm2.0}$ & $\textbf{62.2}_{\pm0.7}$ & $\textbf{79.6}_{\pm0.6}$ & $\textbf{46.6}_{\pm0.6}$ \\
\textbf{MedGrounder} & GMPG & M,P  & 8\,130  & $\underline{90.1}_{\pm1.8}$ & $46.5_{\pm2.1}$ & $77.0_{\pm1.8}$ & $48.3_{\pm2.1}$ & $\textbf{26.0}_{\pm2.7}$ & $\textbf{77.8}_{\pm2.3}$ & $\textbf{37.4}_{\pm2.4}$ & $\underline{62.1}_{\pm1.8}$ & $\underline{79.3}_{\pm2.2}$ & $\underline{46.0}_{\pm2.1}$ \\
\bottomrule
\multicolumn{14}{l}{\scriptsize Fw.\ = Framework. Ann.\ = human-annotated phrases used in training (0 = weak labels only). P@1 denotes P@F1{=}1.} \\
\multicolumn{14}{l}{\scriptsize Datasets: M = MS-CXR, P = PadChest-GR, U = US-Mix. Asterisk (*) = single-box portion only.} \\
\end{tabular}%
}
\end{table*}

% ============================================================
% TABLE 2: MS-CXR
% ============================================================
\begin{table*}[t]
\centering
\caption{\textbf{Grounding performance on the test set of MS-CXR.} Models are grouped into (A)~not trained on MS-CXR and (B)~trained/fine-tuned on MS-CXR.}
\label{tab:mscxr_results}
\resizebox{\textwidth}{!}{%
\begin{tabular}{ll cc ccc ccc ccc}
\toprule
& & & & \multicolumn{3}{c}{\textbf{Single ($n{=}138$)}} & \multicolumn{3}{c}{\textbf{Multi ($n{=}38$)}} & \multicolumn{3}{c}{\textbf{Overall ($n{=}176$)}} \\
\cmidrule(lr){5-7} \cmidrule(lr){8-10} \cmidrule(lr){11-13}
\textbf{Model} & \textbf{Fw.} & \textbf{Data} & \textbf{Ann.}
& $\textbf{P@1}\phantom{_{\pm0.0}}$ & $\textbf{CH-F1}\phantom{_{\pm0.0}}$ & $\textbf{Acc}\phantom{_{\pm0.0}}$
& $\textbf{P@1}\phantom{_{\pm0.0}}$ & $\textbf{CH-F1}\phantom{_{\pm0.0}}$ & $\textbf{Acc}\phantom{_{\pm0.0}}$
& $\textbf{P@1}\phantom{_{\pm0.0}}$ & $\textbf{CH-F1}\phantom{_{\pm0.0}}$ & $\textbf{Acc}\phantom{_{\pm0.0}}$ \\
\midrule
\multicolumn{13}{l}{\textit{(A) Models not trained on MS-CXR}} \\[2pt]
AGPT (pre.)            & MPG  & --    & 0      & $\underline{29.7}\phantom{_{\pm0.0}}$ & $\underline{60.9}\phantom{_{\pm0.0}}$ & $\underline{29.0}\phantom{_{\pm0.0}}$ & $\underline{0.0}\phantom{_{\pm0.0}}$ & $\underline{14.8}\phantom{_{\pm0.0}}$ & $\underline{0.0}\phantom{_{\pm0.0}}$ & $\underline{23.3}\phantom{_{\pm0.0}}$ & $\underline{47.6}\phantom{_{\pm0.0}}$ & $\underline{22.7}\phantom{_{\pm0.0}}$ \\
MedGrounder (pre.)     & GMPG & --    & 0      & $\textbf{43.5}\phantom{_{\pm0.0}}$    & $\textbf{78.7}\phantom{_{\pm0.0}}$    & $\textbf{45.7}\phantom{_{\pm0.0}}$    & $\textbf{13.2}\phantom{_{\pm0.0}}$   & $\textbf{81.5}\phantom{_{\pm0.0}}$    & $\textbf{18.4}\phantom{_{\pm0.0}}$   & $\textbf{36.9}\phantom{_{\pm0.0}}$    & $\textbf{79.9}\phantom{_{\pm0.0}}$    & $\textbf{39.8}\phantom{_{\pm0.0}}$ \\
\midrule
\multicolumn{13}{l}{\textit{(B) Models trained on MS-CXR}} \\[2pt]
TransVG                & MPG  & M*,P* & 3\,993  & $\underline{71.7}\phantom{_{\pm0.0}}$ & $\textbf{94.6}\phantom{_{\pm0.0}}$    & $\underline{71.0}\phantom{_{\pm0.0}}$ & $0.0\phantom{_{\pm0.0}}$             & $59.0\phantom{_{\pm0.0}}$             & $7.9\phantom{_{\pm0.0}}$             & $56.2\phantom{_{\pm0.0}}$             & $83.8\phantom{_{\pm0.0}}$             & $57.4\phantom{_{\pm0.0}}$ \\
MedRPG                 & MPG  & M*    & 624    & $\underline{71.7}\phantom{_{\pm0.0}}$ & $\underline{93.4}\phantom{_{\pm0.0}}$ & $\textbf{73.9}\phantom{_{\pm0.0}}$    & $0.0\phantom{_{\pm0.0}}$             & $54.0\phantom{_{\pm0.0}}$             & $2.6\phantom{_{\pm0.0}}$             & $56.2\phantom{_{\pm0.0}}$             & $81.6\phantom{_{\pm0.0}}$             & $58.5\phantom{_{\pm0.0}}$ \\
AGPT                   & MPG  & M*    & 624    & $\textbf{73.2}\phantom{_{\pm0.0}}$    & $90.8\phantom{_{\pm0.0}}$             & $\textbf{73.9}\phantom{_{\pm0.0}}$    & $0.0\phantom{_{\pm0.0}}$             & $50.7\phantom{_{\pm0.0}}$             & $5.3\phantom{_{\pm0.0}}$             & $\textbf{57.4}\phantom{_{\pm0.0}}$    & $78.7\phantom{_{\pm0.0}}$             & $\underline{59.1}\phantom{_{\pm0.0}}$ \\
MAIRA-2 PG              & GRG  & U,M,P & 73\,043 & $55.8\phantom{_{\pm0.0}}$             & $91.2\phantom{_{\pm0.0}}$             & $55.1\phantom{_{\pm0.0}}$             & $\underline{26.3}\phantom{_{\pm0.0}}$ & $\underline{81.5}\phantom{_{\pm0.0}}$ & $\underline{36.8}\phantom{_{\pm0.0}}$ & $49.4\phantom{_{\pm0.0}}$             & $\underline{88.0}\phantom{_{\pm0.0}}$ & $51.1\phantom{_{\pm0.0}}$ \\
MAIRA-2 GR              & GRG  & U,M,P & 73\,043 & $50.7\phantom{_{\pm0.0}}$             & $86.6\phantom{_{\pm0.0}}$             & $52.2\phantom{_{\pm0.0}}$             & $0.0\phantom{_{\pm0.0}}$             & $18.9\phantom{_{\pm0.0}}$             & $15.8\phantom{_{\pm0.0}}$             & $39.8\phantom{_{\pm0.0}}$             & $77.0\phantom{_{\pm0.0}}$             & $44.3\phantom{_{\pm0.0}}$ \\
\textbf{MedGrounder}   & GMPG & M     & 815    & $65.2_{\pm2.6}$  & $89.1_{\pm1.2}$  & $65.7_{\pm2.4}$  & $\textbf{27.4}_{\pm4.0}$ & $\textbf{83.7}_{\pm1.2}$ & $\textbf{41.1}_{\pm3.0}$ & $\underline{57.0}_{\pm2.1}$ & $\textbf{88.3}_{\pm0.9}$ & $\textbf{60.3}_{\pm2.1}$ \\
\textbf{MedGrounder}   & GMPG & M,P   & 8\,130  & $58.8_{\pm2.2}$  & $87.3_{\pm0.6}$  & $59.7_{\pm2.3}$  & $22.1_{\pm3.0}$ & $80.3_{\pm5.2}$ & $31.1_{\pm4.3}$ & $50.9_{\pm2.1}$ & $85.9_{\pm2.0}$ & $53.5_{\pm1.5}$ \\
\bottomrule
\multicolumn{13}{l}{\scriptsize Fw.\ = Framework. Ann.\ = human-annotated phrases (0 = weak labels only). P@1 denotes P@F1{=}1.} \\
\multicolumn{13}{l}{\scriptsize Datasets: M = MS-CXR, P = PadChest-GR, U = US-Mix. * = single-box portion only.} \\
\end{tabular}%
}
\end{table*}
\section{Results \& discussion}
\subsection{GMPG evaluation}
\subsubsection{Main results on PadChest-GR and MS-CXR}
Tables~\ref{tab:padchest_results} and~\ref{tab:mscxr_results} summarise results in both zero-shot and fine-tuned settings. Across both datasets, MedGrounder is consistently stronger on the aspect that most distinguishes GMPG from MPG, namely multi-region grounding (Multi).

In the zero-shot setting (Panel A in both tables), MedGrounder achieves the best performance on both PadChest-GR and MS-CXR across all reported metrics. One of the baseline models, AGPT, was pre-trained on anatomy-name grounding, where short anatomical labels (e.g., ``left lower lung'', ``heart'') are aligned to regions, whereas MedGrounder was pre-trained on sentence-level alignments between report sentences and anatomical regions. This richer supervision transfers more effectively to GMPG, which requires grounding radiographic phrases rather than bare anatomy names, and particularly benefits set prediction and confidence-based abstention. This advantage is consistent across both datasets.

After fine-tuning (Panel B in both tables), MPG-style methods remain competitive on single-box phrases, especially on MS-CXR. However, MedGrounder shows its clearest advantage on multi-box phrases. On both datasets, it achieves the best multi-box P@F1=1, and the best overall performance on PadChest-GR. On MS-CXR, MPG-style models are strong on single-box phrases, with AGPT achieving the highest single-box P@F$_1{=}1$ (73.2). In contrast, MedGrounder achieves the best multi-box performance, reaching 27.4 P@F$_1{=}1$, 83.7 CH-F1, and 41.1 mask IoU accuracy (Acc), and also delivers the highest overall CH-F1 (88.3) and Acc (60.3). The reduced single-box performance (e.g., 65.2 vs.\ 73.2 for AGPT on MS-CXR) is possibly because of the harder prediction task: MedGrounder must jointly determine how many boxes to predict, where to place them, and when to suppress all predictions, whereas MPG models only regress a single box per phrase.

Figure~\ref{fig:mpg_gmpg} provides a qualitative comparison of TransVG, MedRPG, MAIRA-2 GR, and MedGrounder on three PadChest-GR examples. The bracketed numbers following sentences indicate the ground-truth box count ([0] for non-groundable findings). For all models, each sentence is processed independently. Since this dataset was used to train MAIRA-2 in its grounded report generation mode, we use that same mode for a fair comparison. Each panel reports N-Acc, Prec@F1, and Mask IoU scores. MPG models are constrained to one box per sentence, while GRG and GMPG models can predict multiple boxes, better matching the ground truth. For non-groundable sentences, only MAIRA-2 GR and MedGrounder correctly abstain from producing boxes. When both models abstain correctly, MedGrounder consistently achieves better localisation accuracy on the groundable findings.

\subsubsection{Impact of pre-training and fine-tuning stage}
\begin{table}[!htbp]
\centering
\scriptsize
\setlength{\tabcolsep}{0.75pt}
\caption{
MedGrounder pre-training and fine-tuning strategy ablation.
}
\begin{tabular}{lcccccccc}
\toprule
& \multicolumn{3}{c}{\textbf{MS-CXR} } & \multicolumn{4}{c}{\textbf{PadChest-GR} } \\
\cmidrule(lr){2-4} \cmidrule(lr){5-8}
\textbf{Training regime} &
\textbf{P@F1=1} & \textbf{CH-F1}  & \textbf{mIoU} &
\textbf{P@F1=1} & \textbf{CH-F1}  & \textbf{mIoU} & \textbf{N-Acc} \\
\midrule
FT PadChest-GR            & 0 & 0 & 8.1 & 41.3 & 1.5 & 8.9 & 99.8\\
FT MS-CXR                  & 0 & 4.7 & 22.0  & 41.3 & 2.9 & 24.9 & \textbf{99.9} \\
Pre-train (Chest ImaGenome) & 20.8 & 68.0 & 31.0 & 6.3 & 39.0 & 27.0 & 0.0\\
Pre-train (GMPG-ImaGenome)  & 36.9   & 79.9 & 42.7 & 45.2 & 58.2 & 32.3 & 86.5\\
Pre-train$\rightarrow$FT PC-GR     & 42.0 & 80.7 & 47.9 & 57.7 & 70.8 & 44.4 &  82.6 \\
Pre-train$\rightarrow$FT MS-CXR    & \textbf{58.5 }& \textbf{88.4} &\textbf{55.8} & 44.0 & 56.5 & 32.2 & 84.7  \\
Pre-train$\rightarrow$FT PC-GR+MS-CXR & 54.0 & 86.7 & 54.1 &\textbf{ 61.3} & \textbf{73.2} & \textbf{46.8} & 85.7 \\
\bottomrule
\end{tabular}

\label{tab:two_stage_ablation}
\vspace{0.25em}
\end{table}
Table~\ref{tab:two_stage_ablation} ablates MedGrounder's two-stage training strategy. All models were initialised from general MDETR checkpoints. We compare three regimes: (1)~fine-tuning directly from general checkpoints on the small expert datasets without pre-training, (2)~pre-training alone on either the original Chest ImaGenome or our cleaned GMPG-ImaGenome, and (3)~pre-training on GMPG-ImaGenome followed by fine-tuning. All pre-training rows use GMPG-ImaGenome unless otherwise noted. Three main findings emerge.

First, weak-to-expert training is necessary for GMPG. Without pre-training, the general checkpoints fine-tuned directly on the small expert datasets achieve poor CH-F1, particularly on MS-CXR, indicating that these datasets alone are insufficient for learning effective grounding.

Second, data cleaning is critical for both localisation and groundability. Pre-training on the original Chest ImaGenome already provides some localisation signal, but it fails completely on non-groundable phrases, with 0.0 N-Acc on PadChest-GR. In contrast, pre-training on the cleaned GMPG-ImaGenome improves zero-shot CH-F1 on both datasets and achieves strong N-Acc, showing the effectiveness of our constructed dataset.

Third, fine-tuning further improves performance beyond zero-shot transfer. The best strategy depends on the target dataset: fine-tuning on MS-CXR alone performs best on MS-CXR, while joint fine-tuning on PadChest-GR and MS-CXR performs best on PadChest-GR. The fact that joint fine-tuning does not improve MS-CXR motivates the disease-level cross-dataset analysis in Table~\ref{tab:disease_cross_dataset}.

\subsubsection{Disease-specific analysis on MS-CXR}

% ===================== TABLE =====================
\begin{table}[t]
\centering
  \caption{\textbf{Disease-specific performance and training data characteristics on MS-CXR.} Performance reported as percentages. Rows ordered by descending P@F1. N: number of training samples;
 Area: mean bounding box area as a percentage of image size; ImaGenome-MS Alignment: mean IoU between MS-CXR boxes and the closest matching anatomy box from GMPG-ImaGenome on the same image, reflecting cross-dataset spatial consistency.}
\label{tab:disease_perf_table}
\footnotesize
\setlength{\tabcolsep}{2pt}
\begin{tabular}{l ccc ccc}
\toprule
& \multicolumn{3}{c}{\textbf{Performance Metrics (\%)}} & \multicolumn{3}{c}{\textbf{Training Data Characteristics}} \\
\cmidrule(lr){2-4} \cmidrule(lr){5-7}
\multirow{2}{*}{\textbf{Disease Name}} & \multirow{2}{*}{\textbf{P@F1=1}} & \multirow{2}{*}{\textbf{CH-F1}} & \multirow{2}{*}{\textbf{mIoU}} & \multirow{2}{*}{\textbf{N}} & \multirow{2}{*}{\textbf{Area}} & \textbf{ImaGenome-MS} \\
& & & & & & \textbf{Alignment} \\
\midrule
Cardiomegaly     & 100.0 & 100.0 & 78.3 & 232 & 17.2 & 61.5 \\
Atelectasis      & 62.5  & 96.0  & 51.0 & 72  & 5.8  & 55.7 \\
Pleural Effusion & 50.0  & 95.0  & 51.4 & 97  & 5.5  & 41.9 \\
Pneumonia        & 46.7  & 86.0  & 46.8 & 163 & 7.5  & 51.3 \\
Consolidation    & 40.0  & 90.9  & 47.6 & 138 & 7.3  & 45.3 \\
Pneumothorax     & 38.9  & 74.7  & 45.3 & 182 & 5.0  & 48.3 \\
Edema            & 25.0  & 82.8  & 41.6 & 61  & 11.5 & 49.0 \\
Lung Opacity     & 16.7  & 73.0  & 34.4 & 75  & 6.4  & 35.0 \\
\bottomrule
\end{tabular}
\end{table}

Table~\ref{tab:disease_perf_table} shows large variation across diseases, with P@F$_1{=}1$ ranging from 100.0 for cardiomegaly to 16.7 for lung opacity. CH-F1 remains relatively high throughout, suggesting that the model often identifies the correct general region even when exact localisation is imperfect. To understand what drives this variation, we examine three training data characteristics. Neither the number of pre-training samples (N) nor the average box area correlates strongly with performance: pneumothorax has the second-largest N (182) but ranks among the weaker categories, while atelectasis has the fewest samples (72) yet achieves the second-best P@F$_1{=}1$. In contrast, ImaGenome-MS Alignment, defined as the mean IoU between each MS-CXR box and its closest anatomy box from Chest ImaGenome, tracks performance much more consistently. This is expected: since our pre-training relies on anatomical region supervision, diseases whose ground-truth boxes align well with a single anatomical region receive a stronger and more precise training signal. Cardiomegaly, for instance, maps directly to the cardiac silhouette, whereas pulmonary findings such as lung opacity or pneumothorax overlap with anatomical regions to varying degrees, and those with lower overlap consistently yield weaker grounding performance.

\begin{table}[t]
\centering
  \caption{\textbf{Disease-specific cross-dataset analysis on MS-CXR.} For each disease in MS-CXR, the left
   half compares P@F1 between two models: FT\textsubscript{MS} (fine-tuned on MS-CXR only) and FT\textsubscript{MS+PC} (fine-tuned jointly on MS-CXR and PadChest-GR); $\Delta$ = FT\textsubscript{MS+PC} $-$ FT\textsubscript{MS}. The right half measures whether annotations for the same disease are consistent across the two datasets: N\textsubscript{PC} is the number of matching PadChest-GR samples (--- = disease not present in PadChest-GR), $\Delta$Area is the absolute difference in mean normalised box area (\%) between matched diseases, and $\Delta$Center is the Euclidean distance between mean normalised box centres of the same
  disease across the two datasets.}
\label{tab:disease_cross_dataset}
\footnotesize
\setlength{\tabcolsep}{2pt}
\begin{tabular}{l ccc ccc}
\toprule
& \multicolumn{3}{c}{\textbf{Model Performance (P@F1})} & \multicolumn{3}{c}{\textbf{Dataset Discrepancy}} \\
\cmidrule(lr){2-4} \cmidrule(lr){5-7}
\textbf{MS Diseases} & \textbf{FT\textsubscript{MS}} & \textbf{FT\textsubscript{MS+PC}} & \textbf{$\Delta$} & \textbf{N\textsubscript{PC}} & \textbf{$\Delta$Area} & \textbf{$\Delta$Center} \\
\midrule
Cardiomegaly & 100.0 & 100.0 & 0.0 & 349 & 0.6 & 0.015 \\
Atelectasis & 62.5 & 37.5 & $-$25.0 & 185 & 1.8 & 0.088 \\
Pleural Effusion & 50.0 & 14.3 & $-$35.7 & 148 & 1.2 & 0.086 \\
Pneumonia & 46.7 & 43.3 & $-$3.4 & --- & --- & --- \\
Consolidation & 40.0 & 40.0 & 0.0 & 33 & 0.4 & 0.105 \\
Pneumothorax & 38.9 & 41.7 & +2.8 & 7 & 1.2 & 0.144 \\
Edema & 25.0 & 25.0 & 0.0 & --- & --- & --- \\
Lung Opacity & 16.7 & 8.3 & $-$8.4 & --- & --- & --- \\
\bottomrule
\end{tabular}
\end{table}

\subsubsection{Impact of joint fine-tuning on MS-CXR}
Table~\ref{tab:two_stage_ablation} shows that joint fine-tuning achieves the best PadChest-GR results but hurts MS-CXR compared to fine-tuning on MS-CXR alone. To understand why, Table~\ref{tab:disease_cross_dataset} breaks down MS-CXR results by disease, comparing a model fine-tuned on MS-CXR alone (FT\textsubscript{MS}) against one fine-tuned on both datasets (FT\textsubscript{MS+PC}). For each MS-CXR disease, we further check whether it appears in PadChest-GR, and if so, report the number of matching samples (N\textsubscript{PC}) and two measures of annotation discrepancy: $\Delta$Area (difference in mean normalised box area) and $\Delta$Center (distance between mean normalised box centres).

The impact of joint training varies by disease. Pleural effusion and atelectasis suffer the largest drops ($-$35.7 and $-$25.0). Both have a large number of PadChest-GR samples (148 and 185) with high $\Delta$Center (0.086 and 0.088), meaning the two datasets annotate these diseases in different locations. Cardiomegaly also has many PadChest-GR samples (349) but its annotations are closely aligned ($\Delta$Center~=~0.015), so performance is unaffected. Consolidation and pneumothorax have high $\Delta$Center (0.105 and 0.144) but too few PadChest-GR samples (33 and 7) to matter. Diseases absent from PadChest-GR (pneumonia, edema, lung opacity) show little change. In summary, adding PadChest-GR hurts MS-CXR precisely for diseases where PadChest-GR provides many samples with different annotation conventions, overriding what the model learns from MS-CXR.

\subsubsection{Qualitative analysis on MS-CXR and PadChest-GR}
\begin{figure*}[!t]
    \centering
    \includegraphics[width=1\linewidth]{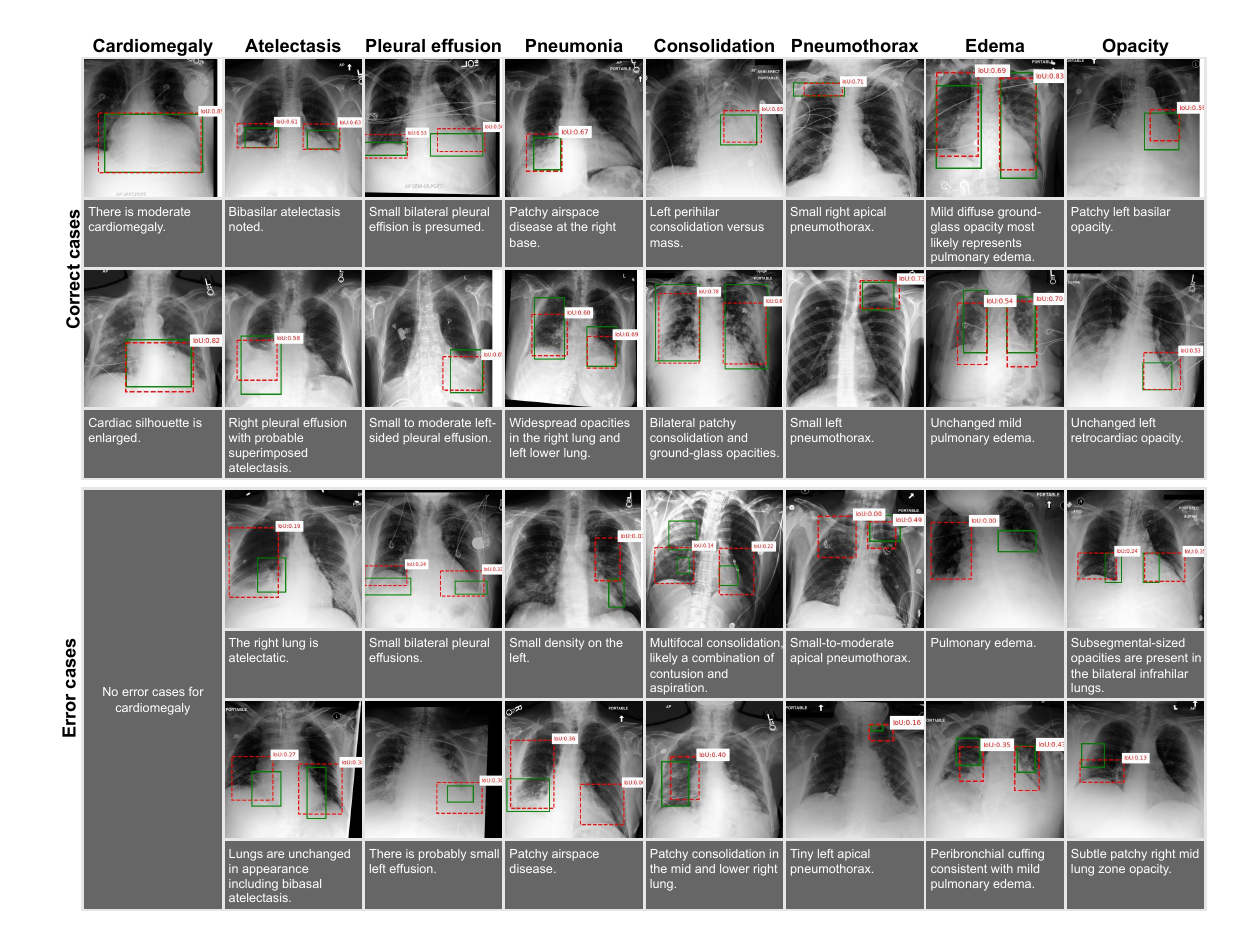}
    \caption{\textbf{Visualisation result of our model on MS-CXR across eight findings.} Ground truth boxes are green, model predictions are red (dashed) with IoU shown. Top: correct cases. Bottom: error cases for the same disease; no cardiomegaly error example appears.}
    \label{fig:gmpg_ms}
\end{figure*}

\begin{figure*}[!t]
    \centering
    \includegraphics[width=1\linewidth]{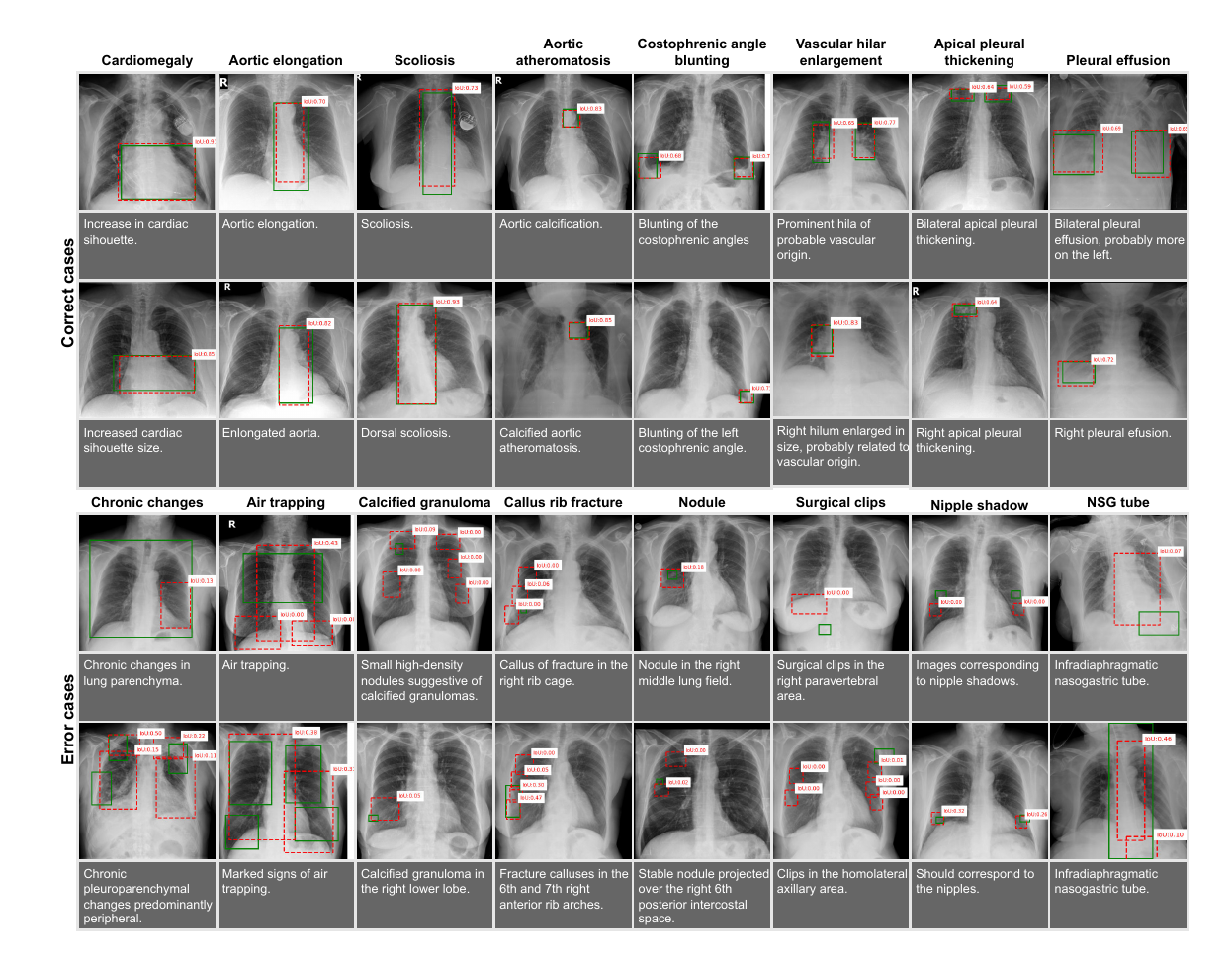}
    \caption{\textbf{Visualisation result of our model on PadChest-GR across sixteen findings.} The top two rows show eight disease categories with strong phrase grounding results. The bottom two rows show eight categories with weaker performance. Ground-truth regions are green; predicted boxes are red (dashed) with IoU indicated.}
    \label{fig:gmpg_pc}
\end{figure*}

Figures~\ref{fig:gmpg_ms} and~\ref{fig:gmpg_pc} visualise MedGrounder's predictions on MS-CXR and PadChest-GR respectively. Across both datasets, performance is strongest for anatomically constrained findings. Phrases with explicit spatial cues (e.g., ``bibasilar'', ``left lower lung'') or findings tied to well-defined structures (e.g., ``cardiomegaly'', ``aortic elongation'') are consistently localised correctly, consistent with the high ImaGenome-MS Alignment observed for these categories in Table~\ref{tab:disease_perf_table}.

Three failure patterns emerge. First, some diseases have no clear boundaries on chest radiographs (e.g., pleural effusion, edema), making precise localisation difficult even for human annotators. Second, some phrases do not specify laterality (e.g., ``small to moderate apical pneumothorax''), so the model must guess location from the image alone. Third, some findings have low anatomical alignment (e.g., calcified granuloma, nodule, rib fracture) and no consistent anatomical anchor. The model responds by producing multiple low-confidence boxes across candidate regions.

\subsection{MedGrounder for grounded report generation}

\begin{table}[!htbp]
\centering
\caption{\textbf{RadFact Evaluation on PadChest-GR.} We report metrics on the full set ($n=915$) and a non-overlapping subset ($n=38$). CXRMate-RRG24 is evaluated only on the subset to avoid training data overlap.}
\label{tab:padchest_radfact_only}
\footnotesize
\setlength{\tabcolsep}{2pt}
\begin{tabular}{l l c c c}
\toprule
\textbf{Model} & \textbf{Set} & \textbf{Logical} & \textbf{Grounding} & \textbf{Spatial}\\
\midrule
MAIRA-2 GR & Full & \textbf{53.89} & \textbf{77.63} & 27.92 \\
MAIRA-2 FG + \textbf{MedGrounder} & Full & 53.69 & 75.41 & \textbf{28.17} \\
\midrule
MAIRA-2 GR & Subset & \textbf{70.91} & 89.89 & 52.03 \\
MAIRA-2 FG + \textbf{MedGrounder} & Subset & 69.21 & 95.45 & \textbf{53.52} \\
CXRMate-RRG24 + \textbf{MedGrounder} & Subset & 65.75 & \textbf{97.14} & 41.58 \\
\bottomrule
\end{tabular}
\end{table}

\begin{table}[!htbp]
\centering
\caption{\textbf{Inference efficiency on PadChest-GR.} Latency measured on 50 images.}
\label{tab:latency}
\footnotesize
\begin{tabular}{llccc}
\toprule
& \textbf{Model} & \textbf{Params} & \textbf{ms/img} & \textbf{Speedup} \\
\midrule
\textbf{End-to-end} & MAIRA-2 GR & 6.88B & 1229.4 & 1.0$\times$ \\
\midrule
\multirow{3}{*}{\textbf{Pipeline}}
& CXRMate-RRG24 & 207M & 151.6 & \\
& MedGrounder & 155M & 73.0 & \\
\cmidrule(l){2-5}
& \textbf{Total} & \textbf{362M} & \textbf{224.6} & \textbf{5.5$\times$} \\
\bottomrule
\end{tabular}
\end{table}

A key practical advantage of formulating grounding as a standalone task is that MedGrounder can be composed with any existing report generator to produce grounded reports, without retraining the generator or requiring image--report--box triplets. The integration is straightforward: given an input image, a report generator first produces a free-text report; the report is then split into sentences, and each sentence is passed together with the image to MedGrounder, which returns zero, one, or more scored bounding boxes per sentence. Sentences receiving no boxes above the confidence threshold are marked as non-groundable. The final output is a grounded report in which each finding sentence is accompanied by its associated image regions, as illustrated in Figure~\ref{fig:compare_app} (bottom, Application B).

We evaluated this pipeline by comparing an end-to-end baseline (MAIRA-2 GR) with two modular configurations: MAIRA-2 in text-only finding generation mode (MAIRA-2 FG) paired with MedGrounder, and CXRMate-RRG24 paired with MedGrounder. For MAIRA-2 GR and MAIRA-2 FG + MedGrounder, we report results on the full PadChest-GR test set. Since CXRMate-RRG24 was trained on PadChest, which overlaps with the PadChest-GR test set, we removed the overlapping cases and formed a 38-case subset for fair evaluation. We report results on this subset for all three models.

\subsubsection{Grounded report quality}
Table~\ref{tab:padchest_radfact_only} evaluates the modular pipeline against the end-to-end MAIRA-2 GR using three RadFact metrics, each measuring a different aspect. Logical F1 measures text accuracy and is independent of MedGrounder, which only adds bounding boxes after the report has been generated. MAIRA-2 GR achieves the highest Logical F1 across both sets, reflecting that MAIRA-2 is a stronger report generator than CXRMate-RRG24.
Grounding F1 is computed only on logically correct sentences, isolating grounding quality from text errors. On the subset, MedGrounder achieves 95.45\% (with MAIRA-2 FG) and 97.14\% (with CXRMate-RRG24), indicating that MedGrounder grounds reliably when given accurate input. These results are notable given that MAIRA-2 is trained on US-MIX, a large-scale private dataset with human bounding-box annotations, whereas MedGrounder uses approximately one tenth of the human annotations. Spatial F1 captures the combined effect of text and grounding quality across all grounded sentences. MAIRA-2 FG + MedGrounder achieves the highest Spatial F1 on both sets (28.17 on Full, 53.52 on Subset). Since Grounding F1 is already high, the main bottleneck for Spatial F1 is report generation quality. This means that pairing MedGrounder with a better report generator can directly improve end-to-end performance without retraining the grounding model.

Overall, the modular approach is a practical alternative to end-to-end GRG: it achieves competitive grounding quality while offering the flexibility to pair with any report generator.

\subsubsection{Inference efficiency}
Beyond report quality, the modular pipeline offers a substantial efficiency advantage. Table~\ref{tab:latency} compares inference latency between the end-to-end and pipeline approaches. CXRMate-RRG24 + MedGrounder is 5.5$\times$ faster than MAIRA-2 GR (224.6 vs.\ 1229.4\,ms per image) with 19$\times$ fewer parameters (362M vs.\ 6.88B). MedGrounder itself adds only 73.0\,ms, making it a lightweight module that can be appended to any generator with minimal overhead. This efficiency gain stems from the modular design: rather than requiring a single large model to jointly generate text and localise findings, the pipeline delegates each subtask to a smaller, purpose-built model.

\subsection{Ablation studies}
\subsubsection{Impact of post-processing}

\begin{figure}
    \centering
    \includegraphics[width=1.0\linewidth]{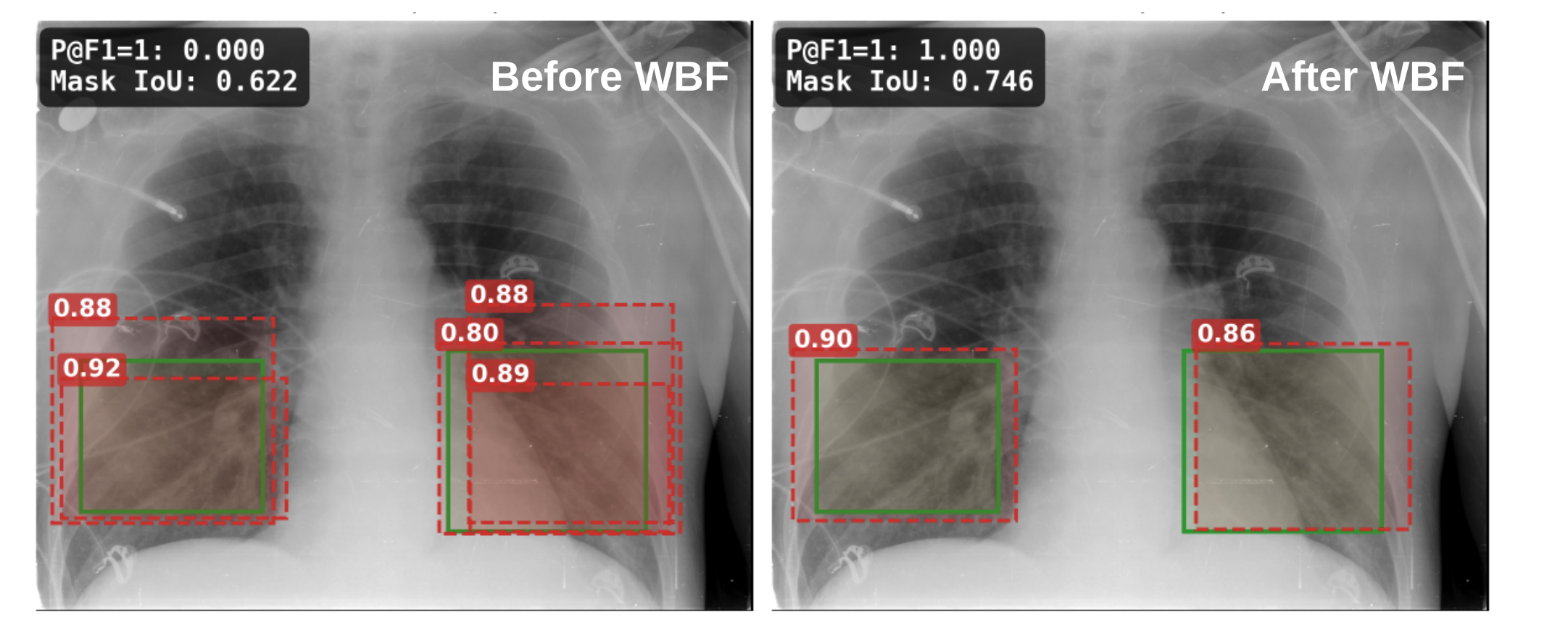}
    \caption{\textbf{Effect of WBF post-processing.} WBF fuses redundant predictions (left) into consolidated boxes (right), improving Mask IoU from 0.622 to 0.746. Green: ground truth; red: predictions with confidence scores.}
    \label{fig:wbf_compare}
\end{figure}

\begin{table}[!htbp]
\centering
\caption{MedGrounder post-processing ablation: weighted box fusion (WBF) enabled vs.\ disabled.}
\label{tab:ablation_postproc_correct}
\scriptsize
\setlength{\tabcolsep}{1.25pt}
\begin{tabular}{lccccccc}
\toprule
& \multicolumn{3}{c}{\textbf{MS-CXR}} & \multicolumn{4}{c}{\textbf{PadChest-GR}} \\
\cmidrule(lr){2-4}\cmidrule(lr){5-8}
\textbf{Post-processing} & \textbf{P@F1=1} & \textbf{CH-F1} & \textbf{mIoU} & \textbf{P@F1=1} & \textbf{CH-F1} & \textbf{mIoU} & \textbf{N-Acc} \\
\midrule
without WBF (baseline) & 44.3 &\textbf{ 86.8} & \textbf{55.1} & 57.9 & \textbf{75.6} & \textbf{48.5} & \textbf{90.1} \\
with WBF & \textbf{51.7} & 85.2 & 54.5 & \textbf{62.9} & 74.5 & 47.3 & \textbf{90.1} \\
\bottomrule
\end{tabular}
\end{table}
Applying Weighted Box Fusion (WBF) with a 0.1 IoU threshold to merge redundant detections further improves performance, raising P@F1=1 by +7.4 on MS-CXR and +5.0 on PadChest-GR (Table~\ref{tab:ablation_postproc_correct}). An example is shown in Figure~\ref{fig:wbf_compare}.

\subsubsection{Impact of in-domain language encoder}

\begin{table}[!htbp]
\centering
\caption{MedGrounder language encoder ablation.}
\label{tab:ablation_lang_all_plus_deltas}
\scriptsize
\setlength{\tabcolsep}{1pt}
\begin{tabular}{lccccccc}
\toprule
& \multicolumn{3}{c}{\textbf{MS-CXR}} & \multicolumn{4}{c}{\textbf{PadChest-GR}} \\
\cmidrule(lr){2-4}\cmidrule(lr){5-8}
\textbf{Model} & \textbf{P@F1=1} & \textbf{CH-F1} & \textbf{mIoU} & \textbf{P@F1=1} & \textbf{CH-F1} & \textbf{mIoU} & \textbf{N-Acc} \\
\midrule
RoBERTa (baseline)     & 29.5 & 78.4 & 38.1 & 44.7 & 57.4 & 19.9 & 85.0 \\
ClinicalBERT & 35.8 & 78.9 & 38.6 & 45.1 & 56.5 & 18.9 & \textbf{86.6} \\
BioClinical ModernBERT   &\textbf{ 36.9} & \textbf{79.9} & \textbf{42.7} & \textbf{45.2} & \textbf{58.2} & \textbf{32.3} & 86.5 \\
\bottomrule
\end{tabular}
\end{table}
Table~\ref{tab:ablation_lang_all_plus_deltas} shows that swapping RoBERTa encoder for a medical-domain model significantly improved performance. The top-performing encoder, BioClinical ModernBERT, boosted the P@F1=1 score by +7.4 on MS-CXR and mIoU by +12.4 on PadChest-GR, confirming that in-domain clinical knowledge is crucial for maximising GMPG performance.

\subsubsection{Comparison with detection frameworks}
Beyond internal design choices, we compare phrase grounding against object detection. Detection takes only an image and predicts boxes for a fixed set of disease classes, while grounding is conditioned on a free-text phrase. We trained DETR~\cite{Carion2020End-to-EndTransformers} (ResNet-101) and Faster R-CNN~\cite{ren2015faster} (ResNet-101+FPN) on the combined MS-CXR and PadChest-GR training data using disease class labels, with the same input resolution and augmentation as MedGrounder. During evaluation, we map each phrase to its disease class and retrieve the detector's predictions for that class.

\begin{table}[h]
\centering
\caption{Comparison with supervised detection frameworks on grounding.}
\label{tab:detection_comparison}
\footnotesize
\setlength{\tabcolsep}{3pt}
\begin{tabular}{lcccccc}
\toprule
& \multicolumn{3}{c}{\textbf{MS-CXR}} & \multicolumn{3}{c}{\textbf{PadChest-GR}} \\
\cmidrule(lr){2-4} \cmidrule(lr){5-7}
\textbf{Method} & \textbf{P@F1=1} & \textbf{CH-F1} & \textbf{Acc} & \textbf{P@F1=1} & \textbf{CH-F1} & \textbf{Acc} \\
\midrule
DETR & 1.1 & 1.8 & 1.1 & 7.4 & 11.6 & 7.4 \\
Faster R-CNN & 30.7 & 42.9 & 31.8 & 17.0 & 35.6 & 18.4 \\
\textbf{MedGrounder} & \textbf{50.9} & \textbf{85.9} & \textbf{53.5} & \textbf{62.1} & \textbf{79.3} & \textbf{46.0} \\
\bottomrule
\end{tabular}
\end{table}

As shown in Table~\ref{tab:detection_comparison}, both detectors substantially underperform MedGrounder. DETR nearly fails, consistent with the limited performance observed when fine-tuning directly from MDETR's general checkpoint (Table~\ref{tab:two_stage_ablation}), suggesting that transformer-based architectures require more training data than available here. Faster R-CNN is more competitive on MS-CXR (8 classes) but struggles on PadChest-GR (154 classes). These results show that disease class labels are insufficient for reliable localisation. The richer semantics in clinical phrases provide disambiguation that detection models cannot access.

\section{Conclusion}
We introduced generalised medical phrase grounding, a clinically aligned formulation that maps each phrase to zero, one, or multiple scored regions. To overcome the scarcity of expert annotations, we constructed GMPG-ImaGenome through an LLM-assisted cleanup of Chest ImaGenome and used it as weak supervision in a two-stage training curriculum. The resulting model, MedGrounder, achieves strong zero-shot transfer, state-of-the-art overall performance on PadChest-GR and competitive results on MS-CXR, with the largest gains on multi-region phrases. By composing MedGrounder with existing report generators, we further showed that grounded report generation can be achieved modularly, without end-to-end training on expensive image--report--box triplets.

Several limitations remain. First, MedGrounder performs well on anatomically constrained findings (e.g., cardiomegaly) but struggles on pulmonary findings with low alignment with anatomical regions (e.g., nodules), and single-box performance remains comparable to rather than exceeding existing MPG methods. Second, both PadChest-GR and MS-CXR contain short, focused annotations (5--6 words), whereas real-world report sentences are longer and more complex (9.2 words in MIMIC-CXR). No phrase grounding dataset with full-length report sentences and bounding box annotations currently exists, making it difficult to evaluate GMPG in a fully realistic clinical setting. One practical direction to address the first limitation is to incorporate existing object-detection datasets that provide accurate annotations for low alignment with anatomical regions. Although these datasets lack sentence-level text, disease names can serve as grounding phrases or be expanded into short descriptions to align with the GMPG setting.

Despite these limitations, MedGrounder opens two practical directions. First, as AI systems increasingly adopt agentic frameworks that orchestrate domain-specific tools, MedGrounder can serve as a grounding
tool that agents call to localise findings in chest X-ray images on demand. Second, MedGrounder can be used as an annotation tool to generate weak bounding box supervision for large-scale unannotated report
datasets, producing image--text--box triplets that benefit not only GMPG but any vision-language task requiring localisation.

\bibliographystyle{IEEEtran}
\bibliography{ref}

@inproceedings{ren2015faster,                                                               
    author    = {Ren, Shaoqing and He, Kaiming and Girshick, Ross and Sun, Jian},
    title     = {Faster {R-CNN}: Towards Real-Time Object Detection with Region Proposal
  Networks},                                                                                  
    booktitle = {{NeurIPS}},
    year      = {2015}                                                                        
  }

@inproceedings{Rezatofighi2019GeneralizedMetric,
    author    = {Rezatofighi, Hamid and Tsoi, Nathan and Gwak, JunYoung and Sadeghian, Amir and Reid, Ian and Savarese, Silvio},
    title     = {Generalized Intersection Over Union: A Metric and a Loss for Bounding Box Regression},                           
    booktitle = {{CVPR}},
    year      = {2019},             
    pages     = {658--666},          
    doi       = {10.1109/CVPR.2019.00075}                  
  }

@article{Kuhn1955Hungarian,
  author  = {Harold W. Kuhn},
  title   = {The {Hungarian} Method for the Assignment Problem},
  journal = {Naval Research Logistics Quarterly},
  year    = {1955},
  volume  = {2},
  number  = {1--2},
  pages   = {83--97},
  doi     = {10.1002/nav.3800020109}
}

@techreport{bannur2024maira2,
  author      = {Bannur, Shruthi and Bouzid, Kenza and Coelho de Castro, Daniel and Schwaighofer, Anton and Bond-Taylor, Sam and Ilse, Maximilian and P\'{e}rez-Garc\'{i}a, Fernando and Salvatelli, Valentina and Sharma, Harshita and Meissen, Felix and Ranjit, Mercy and Srivastav, Shaury and Gong, Julia and Falck, Fabian and Oktay, Ozan and Thieme, Anja and Lungren, Matthew P and Wetscherek, Maria Teodora and Alvarez-Valle, Javier and Hyland, Stephanie},
  title       = {{MAIRA}-2: Grounded Radiology Report Generation},
  institution = {Microsoft},
  year        = {2024}
}

@inproceedings{10.1007/978-3-031-43990-2_35,
  author    = {Chen, Zhihao and Zhou, Yang and Tran, Anh and Zhao, Junting and Wan, Liang and Ooi, Gideon Su Kai and Cheng, Lionel Tim-Ee and Thng, Choon Hua and Xu, Xinxing and Liu, Yong and Fu, Huazhu},
  title     = {Medical Phrase Grounding with Region-Phrase Context Contrastive Alignment},
  booktitle = {{MICCAI}},
  year      = {2023},
  pages     = {371--381}
}

@inproceedings{10204115,
  author    = {Bannur, Shruthi and Hyland, Stephanie and Liu, Qianchu and Perez-Garcia, Fernando and Ilse, Maximilian and Castro, Daniel C. and Boecking, Benedikt and Sharma, Harshita and Bouzid, Kenza and Thieme, Anja and Schwaighofer, Anton and Wetscherek, Maria and Lungren, Matthew P. and Nori, Aditya and Alvarez-Valle, Javier and Oktay, Ozan},
  booktitle = {{CVPR}},
  title     = {Learning to Exploit Temporal Structure for Biomedical Vision-Language Processing},
  year      = {2023},
  pages     = {15016--15027},
  doi       = {10.1109/CVPR52729.2023.01442}
}

@inproceedings{BioViL_10.1007/978-3-031-20059-5_1,
  author    = {Boecking, Benedikt and Usuyama, Naoto and Bannur, Shruthi and Castro, Daniel C. and Schwaighofer, Anton and Hyland, Stephanie and Wetscherek, Maria and Naumann, Tristan and Nori, Aditya and Alvarez-Valle, Javier and Poon, Hoifung and Oktay, Ozan},
  title     = {Making the Most of Text Semantics to Improve Biomedical Vision–Language Processing},
  year      = {2022},
  doi       = {10.1007/978-3-031-20059-5_1},
  booktitle = {{ECCV}},
  pages     = {1--21}
}

@inproceedings{Carion2020End-to-EndTransformers,
  author    = {Nicolas Carion and Francisco Massa and Gabriel Synnaeve and Nicolas Usunier and Alexander Kirillov and Sergey Zagoruyko},
  title     = {End-to-End Object Detection with Transformers},
  booktitle = {{ECCV}},
  year      = {2020},
  pages     = {213--229},
  doi       = {10.1007/978-3-030-58452-8_13}
}

@inproceedings{ChEX_10.1007/978-3-031-72664-4_6,
  author    = {M\"{u}ller, Philip and Kaissis, Georgios and Rueckert, Daniel},
  title     = {{ChEX}: Interactive Localization and Region Description in Chest {X}-Rays},
  year      = {2024},
  doi       = {10.1007/978-3-031-72664-4_6},
  booktitle = {{ECCV}},
  pages     = {92--111}
}

@inproceedings{DBLP:conf/iclr/LoshchilovH19,
  author    = {Ilya Loshchilov and Frank Hutter},
  title     = {Decoupled Weight Decay Regularization},
  booktitle = {{ICLR}},
  year      = {2017}
}

@article{de_Castro_2025,
  title   = {{PadChest-GR}: A Bilingual Chest {X}-Ray Dataset for Grounded Radiology Report Generation},
  volume  = {2},
  doi     = {10.1056/aidbp2401120},
  number  = {7},
  journal = {{NEJM AI}},
  author  = {de Castro, Daniel Coelho and Bustos, Aurelia and Bannur, Shruthi and Hyland, Stephanie L. and Bouzid, Kenza and Wetscherek, Maria Teodora and S\'{a}nchez-Valverde, Maria Dolores and Jaques-P\'{e}rez, Lara and P\'{e}rez-Rodr\'{i}guez, Lourdes and Takeda, Kenji and Salinas-Serrano, Jos\'{e} Mar\'{i}a and Alvarez-Valle, Javier and Galant-Herrero, Joaqu\'{i}n and Pertusa, Antonio},
  year    = {2025},
  month   = {6}
}

@inproceedings{GLoRIA_Huang_2021_ICCV,
  author    = {Huang, Shih-Cheng and Shen, Liyue and Lungren, Matthew P. and Yeung, Serena},
  title     = {{GLoRIA}: A Multimodal Global-Local Representation Learning Framework for Label-Efficient Medical Image Recognition},
  booktitle = {{ICCV}},
  year      = {2021},
  pages     = {3942--3951}
}

@inproceedings{He2016DeepResidual,
  author    = {Kaiming He and Xiangyu Zhang and Shaoqing Ren and Jian Sun},
  title     = {Deep Residual Learning for Image Recognition},
  booktitle = {{CVPR}},
  year      = {2016},
  pages     = {770--778},
  doi       = {10.1109/CVPR.2016.90}
}

@article{He2023GREC,
  title   = {{GREC}: Generalized Referring Expression Comprehension},
  author  = {Shuting He and Henghui Ding and Chang Liu and Xudong Jiang},
  journal = {{arXiv}:2308.16182 [cs.LG]},
  year    = {2023},
  doi     = {10.48550/arXiv.2308.16182}
}

@inproceedings{Hemanthage2024RECANTFormer,
  author    = {Bhathiya Hemanthage and Hakan Bilen and Phil Bartie and Christian Dondrup and Oliver Lemon},
  title     = {{RECANTFormer}: Referring Expression Comprehension with Varying Numbers of Targets},
  booktitle = {{EMNLP}},
  year      = {2024},
  pages     = {21784--21798},
  doi       = {10.18653/v1/2024.emnlp-main.1214}
}

@inproceedings{Ichinose_2023,
  author    = {Akimichi Ichinose and Taro Hatsutani and Keigo Nakamura and Yoshiro Kitamura and Satoshi Iizuka and Edgar Simo-Serra and Shoji Kido and Noriyuki Tomiyama},
  title     = {Visual Grounding of Whole Radiology Reports for {3D} {CT} Images},
  booktitle = {{MICCAI}},
  year      = {2023},
  pages     = {611--621},
  doi       = {10.1007/978-3-031-43904-9_59}
}

@article{johnson2019mimiccxr,
  title   = {{MIMIC-CXR}, a de-identified publicly available database of chest radiographs with free-text reports},
  author  = {Johnson, Alistair EW and Pollard, Tom J and Berkowitz, Seth J and Greenbaum, Nathan R and Lungren, Matthew P and Deng, Chih-ying and Mark, Roger G and Horng, Steven},
  journal = {Scientific Data},
  volume  = {6},
  number  = {1},
  pages   = {317},
  year    = {2019},
  doi     = {10.1038/s41597-019-0322-0}
}

@inproceedings{Kamath2021MDETRUnderstanding,
  author    = {Aishwarya Kamath and Mannat Singh and Yann LeCun and Gabriel Synnaeve and Ishan Misra and Nicolas Carion},
  title     = {{MDETR} -- Modulated Detection for End-to-End Multi-Modal Understanding},
  booktitle = {{ICCV}},
  year      = {2021},
  pages     = {1760--1770},
  doi       = {10.1109/ICCV48922.2021.00180}
}

@inproceedings{kazemzadeh-etal-2014-referitgame,
  title     = {{ReferItGame}: Referring to Objects in Photographs of Natural Scenes},
  author    = {Kazemzadeh, Sahar and Ordonez, Vicente and Matten, Mark and Berg, Tamara},
  booktitle = {{EMNLP}},
  year      = {2014},
  doi       = {10.3115/v1/D14-1086},
  pages     = {787--798}
}

@article{krishna2017visual,
  title   = {{Visual Genome}: Connecting Language and Vision Using Crowdsourced Dense Image Annotations},
  author  = {Krishna, Ranjay and Zhu, Yuke and Groth, Oliver and Johnson, Justin and Hata, Kenji and Kravitz, Joshua and Chen, Stephanie and Kalantidis, Yannis and Li, Li-Jia and Shamma, David A. and Bernstein, Michael S. and Fei-Fei, Li},
  journal = {International Journal of Computer Vision},
  volume  = {123},
  number  = {1},
  pages   = {32--73},
  year    = {2017}
}

@inproceedings{Liu_2021_CVPR,
  author    = {Liu, Fenglin and Wu, Xian and Ge, Shen and Fan, Wei and Zou, Yuexian},
  title     = {Exploring and Distilling Posterior and Prior Knowledge for Radiology Report Generation},
  booktitle = {{CVPR}},
  year      = {2021},
  pages     = {13753--13762}
}

@inproceedings{luo-etal-2025-vividmed,
  title     = {{VividMed}: Vision Language Model with Versatile Visual Grounding for Medicine},
  author    = {Luo, Lingxiao and Tang, Bingda and Chen, Xuanzhong and Han, Rong and Chen, Ting},
  booktitle = {{NAACL}},
  year      = {2025},
  doi       = {10.18653/v1/2025.naacl-long.89},
  pages     = {1800--1821}
}

@misc{meta2024llama3blog,
  author       = {{Meta AI}},
  title        = {Introducing {Meta Llama} 3},
  year         = {2024},
  howpublished = {\url{https://ai.meta.com/blog/meta-llama-3/}}
}

@inproceedings{nicolson-etal-2025-impact,
  title     = {The Impact of Auxiliary Patient Data on Automated Chest {X}-Ray Report Generation and How to Incorporate It},
  author    = {Nicolson, Aaron and Zhuang, Shengyao and Dowling, Jason and Koopman, Bevan},
  booktitle = {{ACL}},
  year      = {2025},
  doi       = {10.18653/v1/2025.acl-long.9},
  pages     = {177--203}
}

@inproceedings{nicolson2024ehealthcsirorrg24entropyaugmented,
  author    = {Aaron Nicolson and Jinghui Liu and Jason Dowling and Anthony Nguyen and Bevan Koopman},
  title     = {e-{Health} {CSIRO} at {RRG}24: Entropy-Augmented Self-Critical Sequence Training for Radiology Report Generation},
  booktitle = {Proceedings of the 23rd Workshop on Biomedical Natural Language Processing},
  year      = {2024},
  pages     = {99--104},
  doi       = {10.18653/v1/2024.bionlp-1.8}
}

@article{NICOLSON2023102633,
  title   = {Improving chest {X}-ray report generation by leveraging warm starting},
  journal = {Artificial Intelligence in Medicine},
  volume  = {144},
  pages   = {102633},
  year    = {2023},
  issn    = {0933-3657},
  doi     = {https://doi.org/10.1016/j.artmed.2023.102633},
  author  = {Aaron Nicolson and Jason Dowling and Bevan Koopman}
}

@article{NICOLSON2024101585,
  title   = {Longitudinal data and a semantic similarity reward for chest {X}-ray report generation},
  journal = {Informatics in Medicine Unlocked},
  volume  = {50},
  pages   = {101585},
  year    = {2024},
  doi     = {https://doi.org/10.1016/j.imu.2024.101585},
  author  = {Aaron Nicolson and Jason Dowling and Douglas Anderson and Bevan Koopman}
}

@inproceedings{nutzel2025generate,
  title     = {Generate to Ground: Multimodal Text Conditioning Boosts Phrase Grounding in Medical Vision-Language Models},
  author    = {N{\"u}tzel, Felix and Dombrowski, Mischa and Kainz, Bernhard},
  booktitle = {Medical Imaging with Deep Learning},
  year      = {2025}
}

@article{sounack2025bioclinicalmodernbertstateoftheartlongcontext,
  title   = {{BioClinical ModernBERT}: A State-of-the-Art Long-Context Encoder for Biomedical and Clinical {NLP}},
  author  = {Thomas Sounack and Joshua Davis and Brigitte Durieux and Antoine Chaffin and Tom J. Pollard and Eric Lehman and Alistair E. W. Johnson and Matthew McDermott and Tristan Naumann and Charlotta Lindvall},
  journal = {{arXiv}:2506.10896 [cs.CL]},
  year    = {2025},
  doi     = {10.48550/arXiv.2506.10896}
}

@inproceedings{TanidaInteractiveGeneration,
  author    = {Tim Tanida and Philip M{\"u}ller and Georgios Kaissis and Daniel Rueckert},
  title     = {Interactive and Explainable Region-guided Radiology Report Generation},
  booktitle = {{CVPR}},
  year      = {2023},
  pages     = {7433--7442},
  doi       = {10.1109/CVPR52729.2023.00718}
}

@article{vilouras2024zero,
  title     = {Zero-shot medical phrase grounding with off-the-shelf diffusion models},
  author    = {Vilouras, Konstantinos and Sanchez, Pedro and O'neil, Alison Q and Tsaftaris, Sotirios A},
  journal   = {{IEEE} Journal of Biomedical and Health Informatics},
  year      = {2024},
  publisher = {IEEE}
}

@article{wu2025towards,
  title   = {Towards generalist foundation model for radiology by leveraging web-scale {2D}\&{3D} medical data},
  author  = {Wu, Chaoyi and Zhang, Xiaoman and Zhang, Ya and Xie, Weidi and Wang, Yanfeng},
  journal = {Nature Communications},
  volume  = {16},
  number  = {1},
  pages   = {7866},
  year    = {2025},
  doi     = {10.1038/s41467-025-62385-7}
}

@article{zhou2025medversageneralistfoundationmodel,
  title   = {{MedVersa}: A Generalist Foundation Model for Medical Image Interpretation},
  author  = {Hong-Yu Zhou and Juli{\'a}n Nicol{\'a}s Acosta and Subathra Adithan and Suvrankar Datta and Eric J. Topol and Pranav Rajpurkar},
  journal = {{arXiv}:2405.07988 [cs.CV]},
  year    = {2024},
  doi     = {10.48550/arXiv.2405.07988}
}

@inproceedings{WuChestReasoning,
  author    = {Wu, Joy T. and Agu, Nkechinyere N. and Lourentzou, Ismini and Sharma, Arjun and Paguio, Joseph A. and Yao, Jasper S. and Dee, Edward C. and Mitchell, William and Kashyap, Satyananda and Giovannini, Andrea and Celi, Leo A. and Moradi, Mehdi},
  title     = {{Chest ImaGenome} Dataset for Clinical Reasoning},
  booktitle = {{NeurIPS}},
  year      = {2021},
  pages     = {1--14}
}

@article{Xiao2024VisualGroundingSurvey,
  title     = {Towards Visual Grounding: A Survey},
  doi       = {10.1109/tpami.2025.3630635},
  journal   = {{IEEE} Transactions on Pattern Analysis and Machine Intelligence},
  publisher = {IEEE},
  author    = {Xiao, Linhui and Yang, Xiaoshan and Lan, Xiangyuan and Wang, Yaowei and Xu, Changsheng},
  year      = {2025},
  pages     = {1--20}
}

@article{Yildirim2024MultimodalRadiology,
  title   = {Multimodal Healthcare {AI}: Identifying and Designing Clinically Relevant Vision-Language Applications for Radiology},
  year    = {2024},
  journal = {Conference on Human Factors in Computing Systems - Proceedings},
  author  = {Yildirim, Nur and Richardson, Hannah and Wetscherek, et al.},
  month   = {2},
  volume  = {22},
  doi     = {10.1145/3613904.3642013}
}

@inproceedings{yu2016modeling,
  title     = {Modeling Context in Referring Expressions},
  author    = {Yu, Licheng and Poirson, Patrick and Yang, Shan and Berg, Alexander C. and Berg, Tamara L.},
  booktitle = {{ECCV}},
  pages     = {69--85},
  year      = {2016}
}

@article{zhang2025development,
  title   = {Development of a large-scale grounded vision language dataset for chest {CT} analysis},
  author  = {Zhang, Xiaoman and Wu, Chaoyi and Zhao, Ziheng and Lei, Jiayu and Tian, Weiwei and Zhang, Ya and Xie, Weidi and Wang, Yanfeng},
  journal = {Scientific Data},
  volume  = {12},
  number  = {1},
  pages   = {1636},
  year    = {2025},
  doi     = {10.1038/s41597-025-05922-9}
}

@inproceedings{Zhang2025Anatomical,
  title     = {Anatomical grounding pre-training for medical phrase grounding},
  author    = {Zhang, Wenjun and Chandra, Shekhar S. and Nicolson, Aaron},
  booktitle = {{ISBI}},
  pages     = {1--5},
  year      = {2025}
}

@article{ZouMedRG:Model,
  title   = {{MedRG}: Medical Report Grounding with Multi-modal Large Language Model},
  author  = {Ke Zou and Yang Bai and Zhihao Chen and Yang Zhou and Yidi Chen and Kai Ren and Meng Wang and Xuedong Yuan and Xiaojing Shen and Huazhu Fu},
  journal = {{arXiv}:2404.06798 [cs.CV]},
  year    = {2024},
  doi     = {10.48550/arXiv.2404.06798}
}

\end{document}